\documentclass{article}

    \PassOptionsToPackage{numbers, compress}{natbib}
    \usepackage[final]{neurips_2022}

\usepackage[utf8]{inputenc} % allow utf-8 input
\usepackage[T1]{fontenc}    % use 8-bit T1 fonts
\usepackage{hyperref}       % hyperlinks
\usepackage{url}            % simple URL typesetting
\usepackage{booktabs}       % professional-quality tables
\usepackage{amsfonts}       % blackboard math symbols
\usepackage{dsfont}
\usepackage{nicefrac}       % compact symbols for 1/2, etc.
\usepackage{microtype}      % microtypography
\usepackage{xcolor}         % colors
\usepackage{comment}

\usepackage{graphicx}       % for table control
\usepackage{caption}
\usepackage{algorithm}
\usepackage{algorithmicx}
\usepackage{algpseudocode}
\usepackage{amsmath}
\usepackage{amsthm}
\usepackage{wrapfig}
\usepackage{amssymb}
\usepackage{multirow}
\usepackage{multicol}
\usepackage{enumitem}
\usepackage{caption}
\usepackage{subcaption}

\newcommand{\ieno}{\textit{i}.\textit{e}.}

\newcommand{\egno}{\textit{e}.\textit{g}.} %there is no space

\newcommand{\etcno}{\textit{etc}} %there is no "."

\newcommand{\ourname}{MLR}

\newcommand{\tcb}{}
\newcommand{\tcr}{}
\newcommand{\tco}{}

\newcommand{\ours}{MLR}

\title{Mask-based Latent Reconstruction for Reinforcement Learning}

\author{%
  Tao Yu$^{1}$\thanks{Equal contribution.} ~\thanks{This work was done when Tao Yu was an intern at Microsoft Research Asia.} ~~
  Zhizheng Zhang$^{2*}$ ~
  Cuiling Lan$^{2}$ ~
  Yan Lu$^{2}$ ~
  Zhibo Chen$^{1}$ \\ $^{1}$University of Science and Technology of China~~~~$^{2}$Microsoft Research Asia \\ 
  \texttt{yutao666@mail.ustc.edu.cn}, \texttt{\{zhizzhang,culan,yanlu\}@microsoft.com} \\ \texttt{chenzhibo@ustc.edu.cn}
  }

\begin{document}

\maketitle

\begin{abstract}
  For deep reinforcement learning (RL) from pixels, learning effective state representations is crucial for achieving high performance. However, in practice, limited experience and high-dimensional \tco{inputs} prevent effective representation learning. To address this, motivated by the success of mask-based modeling in other research fields, we introduce mask-based reconstruction to promote state representation learning in RL. Specifically, we propose a simple yet effective self-supervised method, Mask-based Latent Reconstruction (\ours), to predict complete state representations in the latent space from the observations with spatially and temporally masked pixels. \ours~enables better use of context information when learning state representations to make them more informative, which facilitates \tco{the training of} RL agents. Extensive experiments show that our \ours~significantly improves the sample efficiency in RL and outperforms the state-of-the-art sample-efficient RL methods on multiple continuous and discrete control benchmarks. Our code is available at \href{https://github.com/microsoft/Mask-based-Latent-Reconstruction}{https://github.com/microsoft/Mask-based-Latent-Reconstruction}.
%   For deep reinforcement learning (RL) from pixels, learning effective state representations is crucial for achieving high performance. However, in practice, limited experience and high-dimensional input prevent effective representation learning. To address this, motivated by the success of mask-based modeling in other research fields, we introduce mask-based reconstruction to promote state representation learning in RL. Specifically, we propose a simple yet effective self-supervised method, Mask-based Latent Reconstruction (\ours), to predict the complete state representations in the latent space from the observations with spatially and temporally masked pixels. \ours~enables the better use of context information when learning state representations to make them more informative, which facilitates RL agent training. Extensive experiments show that our \ours~significantly improves the sample efficiency in RL and outperforms the state-of-the-art sample-efficient RL methods on multiple continuous and discrete control benchmarks. Our code is available at \href{https://github.com/microsoft/Mask-based-Latent-Reconstruction}{https://github.com/microsoft/Mask-based-Latent-Reconstruction}.
\end{abstract}

\section{Introduction}
\label{intro}
Learning effective state representations is crucial for reinforcement learning (RL) from visual signals (where a sequence of images is usually the input \tco{of an RL agent}),
% (where a sequence of images is usually the input to an RL network),
such as in DeepMind Control Suite \cite{tassa2018deepmind}, Atari games \cite{bellemare2013arcade}, \etcno. Inspired by the success of \tco{mask-based}
% masked
pretraining in the fields of natural language processing (NLP)~\cite{devlin2018bert,radford2018improving,radford2019language,brown2020language} and computer vision (CV)~\cite{bao2021beit,he2022masked,xie2022simmim}, we make the first endeavor to explore the idea of mask-based \tco{modeling} in RL.
% reconstruction in RL.

% Masked
\tco{Mask-based} pretraining exploits the reconstruction of masked word embeddings or pixels to promote feature learning in NLP or CV \tco{fields}. This is, in fact, not straightforwardly applicable for RL due to the following two reasons. First, RL agents learn \tco{policies} from \tco{the} interactions with environments, where the experienced states vary as the policy network is updated. Intuitively, collecting additional rollouts for pretraining is often costly \tco{especially} in real-world applications. Besides, it is challenging to learn effective state representations without the awareness of the learned policy. Second, visual signals usually have high information \tco{densities},
% density,
which may contain distractions and redundancies for policy learning. Thus, for RL, performing reconstruction in the original (pixel) space is not as necessary as it is in the CV or NLP \tco{fields.}
% domain.

Based on the analysis above, we study the mask-based modeling tailored to vision-based RL. We present Mask-based Latent Reconstruction (MLR), a simple yet effective self-supervised method, to better learn state representations in RL. Contrary to treating mask-based modeling as a pretraining task in the fields of CV and NLP, our proposed MLR is an auxiliary objective optimized together with the policy learning objectives. In this way, the coordination between representation learning and policy learning is considered within a joint training framework.
Apart from this, another key difference compared to vision/language research is that we reconstruct masked pixels in the latent space instead of the input space. We take the state representations (\ieno, features) inferred from original unmasked frames as the reconstruction targets. This effectively reduces unnecessary reconstruction relative to the pixel-level one and further facilitates the coordination between representation learning and policy learning because the state representations are directly optimized.

Consecutive frames are highly correlated. In MLR, we exploit this property to enable the learned state representations to be more informative, predictive and consistent over both spatial and temporal dimensions.
Specifically, we randomly mask a portion of space-time cubes in the input observation (\ieno, video clip) sequence and reconstruct the \tco{feature representations} of the missing contents in the latent space.
% \tcb{reconstruct the features of the} missing contents in the latent space.
%Specifically, we randomly mask a portion of space-time cubes in the input observation (\ieno, video clip) sequence and reconstruct the \tcr{information in the} missing contents in the latent space. 
In this way, similar to the spatial reconstruction for images in \cite{he2022masked,xie2022simmim}, MLR enhances the awareness of the agents \tco{on} 
% of
the global context information of the entire input observations and promotes the state representations to be predictive in both spatial and temporal dimensions. The global predictive information is encouraged to be encoded into each frame-level state representation, achieving better representation learning and further facilitating policy learning.

\tco{We not only propose} an effective mask-based modeling method, \tco{but also conduct} a systematical empirical study for the practices of masking and reconstruction that are as applicable to RL as possible.
% Not only is an effective mask-based modeling method proposed, but we also conduct a systematical empirical study for the practices of masking and reconstruction that are as applicable to RL as possible.
First, we study the influence of masking strategies by comparing spatial masking, temporal masking and spatial-temporal masking. Second, we investigate the differences between masking and reconstructing in the pixel space and in the latent space. Finally, we study how to effectively add reconstruction supervisions in the latent space.

Our contributions are summarized below:

\begin{itemize}[noitemsep,nolistsep,leftmargin=*]
\item We introduce the idea of enhancing representation learning by mask-based \tco{modeling} to RL to improve the sample efficiency. We integrate the mask-based reconstruction into \tco{the training of RL as} an auxiliary objective,
% RL training with an auxiliary objective,
obviating the need for collecting additional rollouts for pretraining and helping the coordination between representation learning and policy learning in RL.
\item We propose Mask-based Latent Reconstruction (MLR), a self-supervised mask-based modeling method to improve the state representations for RL. Tailored to RL, we propose to randomly mask space-time cubes in the pixel space and reconstruct the \tco{information} of
%\tcr{information in the} 
missing contents \textit{in the latent space}. This is shown to be effective for improving the sample efficiency on multiple continuous and discrete control benchmarks (\egno, DeepMind Control Suite \cite{tassa2018deepmind} and Atari games \cite{bellemare2013arcade}).
\item A systematical empirical study is conducted to investigate the good practices of masking and reconstructing operations in \ours~for RL. This demonstrates the effectiveness of our proposed designs in \tco{the proposed} MLR.
\end{itemize}

\section{Related Work}

\subsection{Representation Learning for RL}
Reinforcement learning from visual signals is of high practical value in real-world applications such as robotics, video game AI, \etcno. However, such high-dimensional observations may contain distractions or redundant information, imposing considerable challenges for RL agents to learn effective representations \cite{Shelhamer2017LossII}.
Many prior works address this challenge by taking advantage of self-supervised learning to promote the representation learning of the states in RL. A popular approach is to jointly learn policy learning objectives and auxiliary objectives such as pixel reconstruction \cite{Shelhamer2017LossII, yarats2021improving}, reward prediction \cite{jaderberg2016reinforcement, Shelhamer2017LossII}, bisimulation \cite{zhang2021learning}, dynamics prediction \cite{Shelhamer2017LossII, guo2020bootstrap, lee2020slac, lee2020predictive, schwarzer2021dataefficient, yu2021playvirtual}, and contrastive learning of instance discrimination \cite{laskin2020curl} or (spatial -) temporal discrimination \cite{oord2018representation, anand2019stdim, stooke2020decoupling, zhu2020masked, mazoure2020deep}.
In this line, BYOL~\cite{grill2020byol}-style auxiliary objectives, which
\tco{are often adopted} with data augmentation,
% often cooperate with data augmentation,
show promising performance \cite{schwarzer2021dataefficient,yu2021playvirtual,yarats2021reinforcement,hansen2021generalization,guo2022byol}. More detailed introduction \tco{for}
% to
BYOL and the BYOL-style objectives can be found in Appendix \ref{appendix_byol}.
% Extended introduction to BYOL and the BYOL-style objectives can be found in Appendix \ref{appendix_byol}.
Another feasible way of acquiring good representations is to pretrain the state encoder to learn effective state representations for the original observations before policy learning. It requires additional offline sample collection or early access to the environments \cite{hansen2019fast, stooke2020decoupling, liu2021behavior, liu2021aps, schwarzer2021pretraining}, which is not fully consistent with the principle of sample efficiency in practice. This work aims to design a more effective auxiliary task to improve the learned representations toward sample-efficient RL.

\subsection{Sample-Efficient Reinforcement Learning}
Collecting rollouts from the interaction with the environment is commonly costly, especially in the real world, leaving the sample efficiency of RL algorithms concerned. To improve the sample efficiency of vision-based RL (\ieno, RL from pixel observations), recent works design auxiliary tasks to explicitly improve the learned representations \cite{ yarats2021improving, laskin2020curl,lee2020slac, lee2020predictive, zhu2020masked, liu2021returnbased, schwarzer2021dataefficient, ye2021mastering, yu2021playvirtual} or adopt data augmentation techniques, such as random crop/shift, to improve the diversity of data used for training 
% based on collected samples 
\cite{yarats2021image,laskin2020reinforcement}.
Besides, there are some model-based methods that learn (world) models in the pixel \cite{Kaiser2020Model} or latent space \cite{hafner2019learning, Hafner2020Dream, hafner2020mastering, ye2021mastering}, and perform planning, imagination or policy learning based on the learned models. We focus on the auxiliary task line in this work.

\subsection{Masked Language/Image Modeling}
Masked language modeling (MLM) \cite{devlin2018bert} and its autoregressive variants \cite{radford2018improving, radford2019language, brown2020language} achieve significant success in the NLP field and produce impacts in other domains. MLM masks a portion of word tokens from the input sentence and trains the model to predict the masked tokens, which has been demonstrated to be generally effective in learning language representations for various downstream tasks. For computer vision (CV) tasks, similar to MLM, masked image modeling (MIM) learns representations for images/videos by pretraining the neural networks to reconstruct masked pixels from visible ones. As an early exploration, Context Encoder \cite{pathak2016context} apply this idea to Convolutional Neural Network (CNN) model to train a CNN model with a masked region inpainting task. With the recent popularity of the Transfomer-based architectures, a series of works \cite{chen2020generative, bao2021beit, he2022masked, xie2022simmim, wei2022masked} dust off the idea of MIM and show impressive performance on learning representations for vision tasks. Inspired by MLM and MIM, we explore the mask-based modeling for RL to exploit the high correlation in vision data to improve agents' awareness of global-scope dynamics in learning state representations. Most importantly, \textit{we propose to predict the masked information in the latent space}, instead of the pixel space like the aforementioned MIM works, which better coordinates the representation learning and the policy learning in RL.

% \begin{wrapfigure}{r}{7.cm}
\begin{figure}[t]%[h]
% \vspace{-15pt}
	\begin{center}
		\includegraphics[scale=0.62]{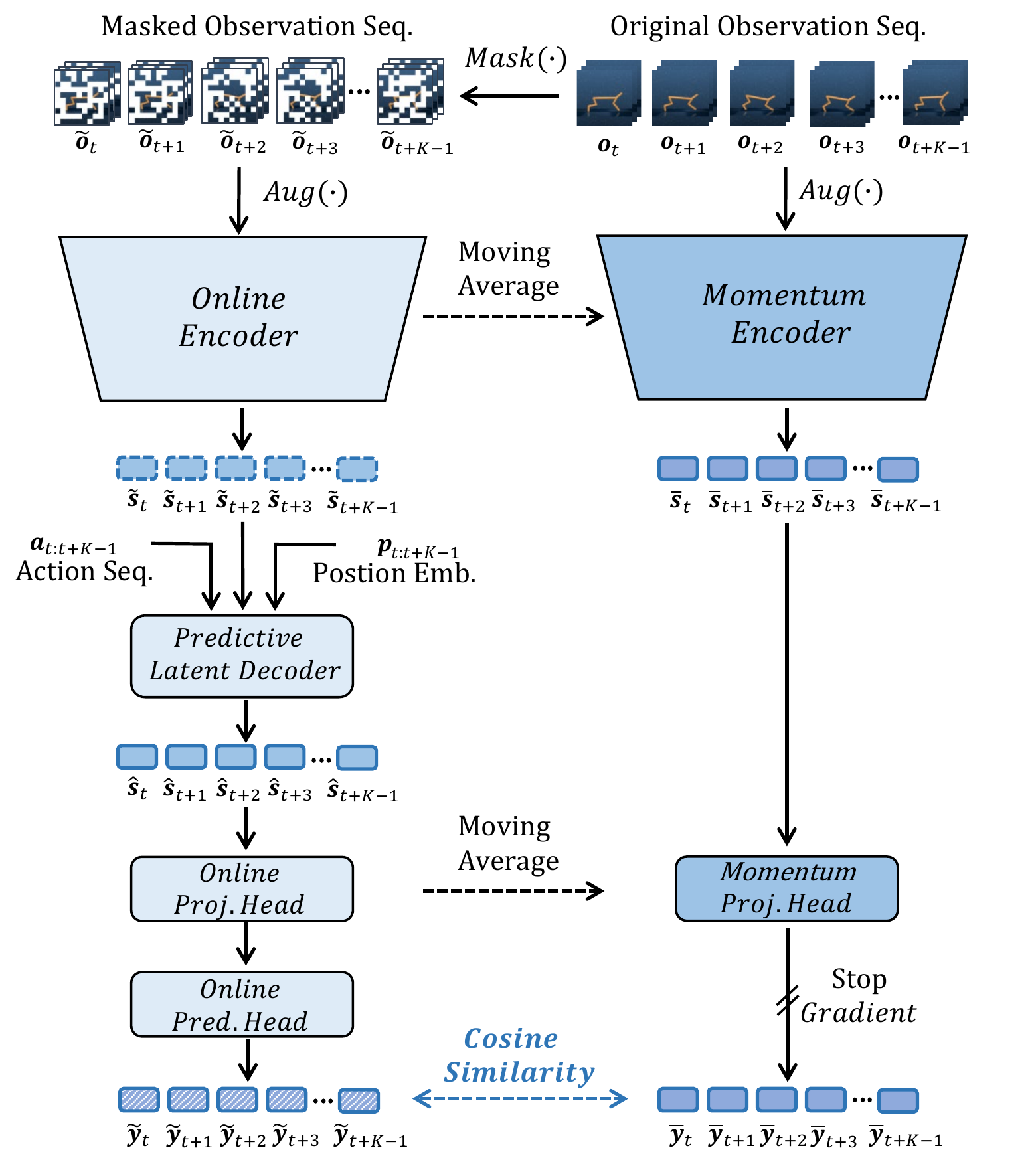} 
	\end{center}
% 	\vspace{-3mm}
	\caption{The framework of the proposed \ours. We perform a random spatial-temporal masking (\ieno, \textit{cube} masking) on the sequence of consecutive observations in the pixel space. The masked observations are encoded to be the latent states through an online encoder. We further introduce a predictive latent decoder to decode/predict the latent states conditioned on the corresponding action sequence and temporal positional embeddings.
	Our method trains the networks to reconstruct the feature representations of the 
	%\tcr{information available in the} 
	missing contents in an appropriate \textit{latent} space using a cosine similarity based distance metric applied between the predicted features of the reconstructed states and the target features inferred from original observations by momentum networks.}
	\label{fig:framework}
\end{figure}
% \end{wrapfigure}

\section{Approach}
\subsection{Background}
Vision-based RL aims to learn policies from pixel observations by interacting with the environment. The learning process corresponds to a partially observable Markov decision process (POMDP) \cite{bellman1957markovian,kaelbling1998planning}, formulated as $(\mathcal{O},\mathcal{A},p,r,\gamma)$, where $\mathcal{O}$, $\mathcal{A}$, $p$, $r$ and $\gamma$ denote the observation space (\egno, pixels), the action space, the transition dynamics $p=Pr(\mathbf{o}_{t+1}|\mathbf{o}_{\le t}, \mathbf{a}_{t})$, the reward function $\mathcal{O}\times \mathcal{A}\to \mathbb{R}$, and the discount factor, respectively.
% Generally, at the timestep $t$, given the observation $\mathbf{o}_{t}$ from the environment, the RL agent responses to this observation by taking an action $\mathbf{a}_{t}$. When the environment receives $\mathbf{a}_{t}$, it will transfer $\mathbf{o}_{t}$ to the next observation $\mathbf{o}_{t+1}$ with a probability $p_{t}$ and return a reward $r_{t}$. 
% Following common practice \cite{mnih2013playing}, several consecutive frames are stacked to construct an observation (which can be seen as a state in practice when converting the POMDP to an MDP) \cite{bellman1957markovian}. 
Following common practices \cite{mnih2013playing}, an observation $\mathbf{o}_{t}$ is constructed by a few RGB frames. The transition dynamics and the reward function can be \tco{formulated}
% \tcb{considered}
as $p_{t}=Pr(\mathbf{o}_{t+1}|\mathbf{o}_{t},\mathbf{a}_{t})$ and $r_{t}=r\left ( \mathbf{o}_{t},\mathbf{a}_{t} \right )$, respectively. 
The objective of RL is to learn a policy $\pi(\mathbf{a}_{t}|\mathbf{o}_{t})$ that maximizes the cumulative discounted return $\mathbb{E}_{\pi}\sum_{t=0}^{\infty }\gamma^{t}r_{t}$, where $\gamma \in [0,1]$.

\subsection{Mask-based Latent Reconstruction}

Mask-based Latent Reconstruction (MLR) is an auxiliary objective to promote representation learning in vision-based RL and is generally applicable \tco{for} different RL algorithms, \egno, Soft Actor-Critic (SAC) and Rainbow \cite{hessel2018rainbow}. The core idea of MLR is to facilitate state representation learning by reconstructing spatially and temporally masked pixels in the latent space. This mechanism enables better use of context information when learning state representations, further enhancing the understanding of RL agents for visual signals. We illustrate the overall framework of MLR in Figure \ref{fig:framework} and elaborate on it below.

\textbf{Framework.} In \ourname, as shown in Figure \ref{fig:framework}, we mask a portion of pixels in the input observation sequence along its spatial and temporal dimensions. We encode the masked sequence and the original sequence from observations to latent states with an encoder and a momentum encoder, respectively. We perform predictive reconstruction from the states corresponding to the masked sequence \tco{while}
% by
taking the states encoded from the original sequence as the target. We add reconstruction supervisions between the prediction results and the targets in the decoded latent space. The processes of \textit{masking}, \textit{encoding}, \textit{decoding} and \textit{reconstruction} are introduced in detail below.

\begin{wrapfigure}{r}{0.5\textwidth}%[h]
% \begin{figure}[t]%[h]
\vspace{-15pt}
	\begin{center}
		\includegraphics[scale=0.85]{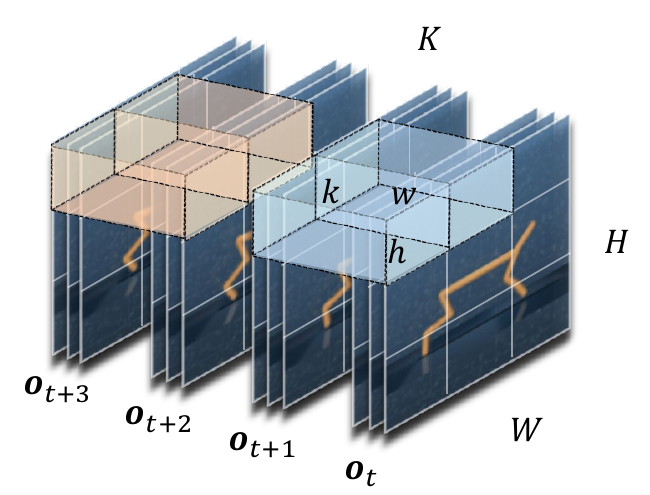} 
	\end{center}
% 	\caption{Illustration of our cube masking. In this example, the observation sequence has $K=4$ timesteps and a spatial size of $H \times W$. We have $h=\frac{1}{3}H$, $w=\frac{1}{3}W$ and $k=\frac{1}{2}K$.}
    \vspace{-10pt}
	\caption{Illustration of our cube masking. We divide the input observation sequence to non-overlapping cubes ($k \times h \times w$). In this example, we have $k=\frac{1}{2}K$, $h=\frac{1}{3}H$ and $w=\frac{1}{3}W$ where the observation sequence has $K=4$ timesteps and a spatial size of $H \times W$.}
	\label{fig:cube}
	\vspace{-15pt}
% \end{figure}
\end{wrapfigure}

\textbf{(i) Masking.}
Given an observation sequence of $K$ timesteps $\tau_K^o=\{\textbf{o}_t, \textbf{o}_{t+1}, \cdots, \textbf{o}_{t+K-1}\}$
% \tcr{(with a slight abuse of notation, for simplicity we define the input to the deep RL networks at each timestep as $\mathbf{o}_t$)}
and each observation containing $n$ RGB frames, all the observations in the sequence are stacked to be a cuboid with the shape of $K\times H\times W$ where $H\times W$ is the spatial size. Note that the actual shape is $K\times H\times W\times D$ where $D=3n$ is the number of channels in each observation. We omit the channel dimension for simplicity. 
As illustrated in Figure \ref{fig:cube}, we divide the cuboid into regular non-overlapping \textit{cubes} with the shape of $k\times h\times w$. We then randomly mask a portion of the cubes following a uniform distribution and obtain a masked observation sequence $\tilde{\tau}_K^o=\{\tilde{\textbf{o}}_t, \tilde{\textbf{o}}_{t+1}, \cdots, \tilde{\textbf{o}}_{t+K-1}\}$. Following \cite{yu2021playvirtual}, we perform stochastic image augmentation $Aug(\cdot)$ (\egno, random crop and intensity) on each masked observation in $\tilde{\tau}_K^o$. The objective of MLR is to predict the state representations of the unmasked observation sequence from the masked one in the latent space.

\textbf{(ii) Encoding.} We adopt two encoders to learn state representations from masked observations and original observations respectively. A CNN-based encoder $f$ is used to encode each masked observation $\tilde{\textbf{o}}_{t+i}$ into its corresponding latent state $\mathbf{\tilde{s}}_{t+i} \in \boldsymbol{\mathbb{R}}^{d}$. After the encoding, we obtain a sequence of the masked latent states $\tilde{\tau}_K^s=\{\mathbf{\tilde{s}}_t, \mathbf{\tilde{s}}_{t+1}, \cdots, \mathbf{\tilde{s}}_{t+K-1}\}$ for masked observations. The parameters of this encoder are updated based on gradient back-propagation in an end-to-end way. We thus call it ``online'' encoder. The state representations inferred from original observations are taken as the targets of subsequently described reconstruction. To make them more robust, inspired by \cite{laskin2020curl,schwarzer2021dataefficient,yu2021playvirtual}, we exploit another encoder for the encoding of original observations. This encoder, called ``momentum'' encoder as in Figure \ref{fig:framework}, has the same architecture as the online encoder, and its parameters are updated by an exponential moving average (EMA) of the online encoder weights $\theta_f$ with the momentum coefficient $m\in [0, 1)$, as formulated below:
\begin{equation}
    \bar{\theta}_f\leftarrow m\bar{\theta}_f+(1-m)\theta_f.
    \label{equ:ema}
\end{equation}
\textbf{(iii) Decoding.}
Similar to the mask-based modeling in the CV field, \egno, \cite{he2022masked,xie2022simmim}, the online encoder in our \ours~predicts the information of the masked contents from the unmasked contents. 
As opposed to pixel-space reconstruction in \cite{he2022masked,xie2022simmim}, \ours~performs the reconstruction in the latent space to better \tco{perform}
% serve
RL policy learning. Moreover, pixels usually have high information 
\tco{densities}
% density
\cite{he2022masked} which may contain distractions and redundancies for the policy learning in RL. Considering the fact that in RL the next state is determined by the current state as well as the action, we propose to utilize both the actions and states as \tco{the inputs} for prediction to reduce the possible ambiguity to enable an RL-tailored mask-based modeling design.
%As opposed to pixel-space reconstruction in \cite{he2022masked,xie2022simmim}, \ours~performs the reconstruction in the latent space since pixels usually have high information density \cite{he2022masked} which may contain distractions and redundancies for the policy learning in RL. Further, we consider the fact in RL that the next state is determined by the current state as well as the action and propose to utilize the actions to reduce the possible ambiguity during prediction as an RL-tailored mask-based modeling design.
To this end, we adopt a transformer-based latent decoder to infer the reconstruction results via a global message passing where the actions and temporal information are exploited. Through this process, the predicted information is ``passed'' to its corresponding state representations.

\begin{wrapfigure}{r}{0.5\textwidth}
% \begin{figure}[t]%[h]
    % \vspace{-10pt}
    % \vspace{-4mm}
	\begin{center}
		\includegraphics[scale=0.525]{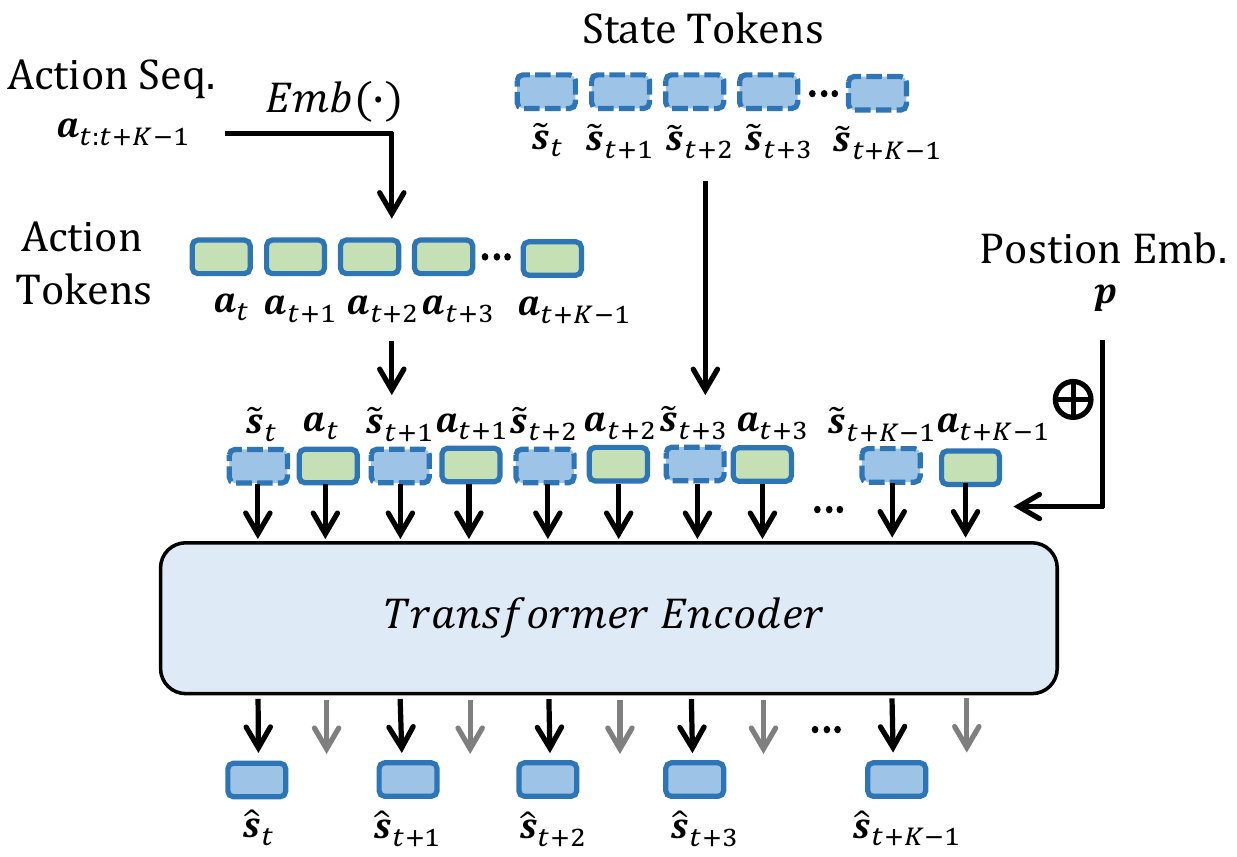} 
	\end{center}
% 	\vspace{-10pt}
	\caption{Illustration of predictive latent decoder. 
% 	The masked latent states . The state and action with   
	}
% 	\vspace{-10pt}
	\label{fig:decoder}
% \end{figure}
\end{wrapfigure}

% Similar to the encoder designs in \cite{he2022masked,xie2022simmim}, the online encoder in our \ours~reconstructs the representations of masked contents based on \tcr{unmasked contents}. The global predictive information has been included in the outputs of the online encoder. To add the reconstruction loss between the state representations inferred from masked and original observations, we need to compare them in a unified latent space. To this end, we leverage a Transformer-based latent decoder to refine the outputs of the online encoder via a global message passing where the actions and temporal information are exploited as the contexts. Through this process, the predicted information is ``passed'' to its corresponding states for adding reconstruction losses.

% Similar to the encoder designs in \cite{he2022masked,xie2022simmim}, the online encoder in our \ours~reconstructs the representations of masked contents based on \tcr{unmasked contents} in an \textit{implicit} way. The global predictive information has been implicitly included in the outputs of the online encoder. To add the reconstruction loss between the state representations inferred from masked and original observations, we need to compare them in a unified latent space. To this end, we leverage a Transformer-based latent decoder to refine the outputs of the online encoder via a global message passing where the actions and temporal information are exploited as the contexts. Through this process, the implicitly predicted information is ``passed'' to its corresponding states for adding reconstruction losses.

As shown in Figure \ref{fig:decoder}, the input tokens of the latent decoder consist of both the masked state sequence $\tilde{\tau}_K^s$ (\ieno, state tokens) and the corresponding action sequence $\tau_K^a=\{\mathbf{a}_t, \mathbf{a}_{t+1}, \cdots, \mathbf{a}_{t+K-1}\}$ (\ieno, action tokens). Each action token is embedded as a feature vector with the same dimension as the stated token using an embedding layer, through an embedding layer $Emb(\cdot)$. Following the common practices in transformer-based models \cite{vaswani2017attention}, we adopt the relative positional embeddings to encode relative temporal positional information into both state and action tokens with an element-wise addition denoted by "$+$" in the following equation (see Appendix \ref{appendix_arch} for more details). Notably, the state and action token at the same timestep $t+i$ share the same positional embedding $\mathbf{p}_{t+i} \in \boldsymbol{\mathbb{R}}^{d}$. 
Thus, the inputs of latent decoder can be mathematically represented as follows:
% As shown in Figure \ref{fig:decoder}, the input tokens of the latent decoder consist of both the masked state sequence $\tilde{\tau}_K^s$ (\ieno, state tokens) and the corresponding action sequence $\tau_K^a=\{\mathbf{a}_t, \mathbf{a}_{t+1}, \cdots, \mathbf{a}_{t+K-1}\}$ (\ieno, action tokens). Each action token is embedded as a feature vector with the same dimension as the stated token using an embedding layer, through an embedding layer $Emb(\cdot)$. We integrate the relative positional embeddings as in \cite{vaswani2017attention} to encode relative temporal positional information into both state and action tokens. Notably, the state and action token at the same timestep $t+i$ share the same positional embedding $\mathbf{p}_{t+i} \in \boldsymbol{\mathbb{R}}^{d}$. Thus, the inputs of latent decoder can be mathematically represented as:
\begin{equation}
    \mathbf{x}=[\mathbf{\tilde{s}}_t, \mathbf{a}_t, \mathbf{\tilde{s}}_{t+1}, \mathbf{a}_{t+1}, \cdots, \mathbf{\tilde{s}}_{t+K-1}, \mathbf{a}_{t+K-1}]+[\mathbf{p}_t, \mathbf{p}_t, \mathbf{p}_{t+1}, \mathbf{p}_{t+1}, \cdots, \mathbf{p}_{t+K-1}, \mathbf{p}_{t+K-1}].
\end{equation}
The input token sequence is \tco{fed into}
% passed through
a Transformer encoder \cite{vaswani2017attention} consisting of $L$ attention layers. Each layer is composed of a Multi-Headed Self-Attention (MSA) layer \cite{vaswani2017attention}, a layer normalisation (LN) \cite{ba2016layer}, and multilayer perceptron (MLP) blocks. The process can be described as follows:
\vspace{-2mm}
\begin{equation}
    \mathbf{z}^l=\boldsymbol{\textrm{MSA}}(\boldsymbol{\textrm{LN}}(\mathbf{x}^l))+\mathbf{x}^l,
    % \vspace{-2mm}
\end{equation}
\begin{equation}
    \mathbf{x}^{l+1}=\boldsymbol{\textrm{MLP}}(\boldsymbol{\textrm{LN}}(\mathbf{z}^l))+\mathbf{z}^l.
    % \vspace{2mm}
\end{equation}
The output tokens of the latent decoder, \ieno, $\hat{\tau}_K^s=\{\mathbf{\hat{s}}_t, \mathbf{\hat{s}}_{t+1}, \cdots, \mathbf{\hat{s}}_{t+K-1}\}$, are the predictive reconstruction results for the latent representations inferred from the original observations. We elaborate on the reconstruction loss between the prediction results and the corresponding targets in the following.

\textbf{(iv) Reconstruction.} Motivated by the success of BYOL \cite{grill2020byol} in self-supervised learning, we use an asymmetric architecture for calculating the distance between the predicted/reconstructed latent states and the target states, similar to \cite{schwarzer2021dataefficient, yu2021playvirtual}. For the outputs of the latent decoder, we use a projection head $g$ and a prediction head $q$ to get the final prediction result $\mathbf{\hat{y}}_{t+i}=q(g(\mathbf{\hat{s}}_{t+i}))$ corresponding to $\mathbf{s}_{t+i}$. For the encoded results of original observations, we use a momentum-updated projection head $\bar{g}$ whose weights are updated with an EMA of the weights of the online projection head. These two projection heads have the same architectures. The outputs of the momentum projection head $\bar{g}$, \ieno, $\mathbf{\bar{y}}_{t+i}=\bar{g}(\mathbf{\bar{s}}_{t+i})$, are the final reconstruction targets. Here, we apply a stop-gradient operation as illustrated in Figure \ref{fig:framework} to avoid model collapse, following \cite{grill2020byol}.

The objective of MLR is to enforce the final prediction result $\mathbf{\hat{y}}_{t+i}$ to be as close as possible to its corresponding target $\mathbf{\bar{y}}_{t+i}$. To achieve this, we design the reconstruction loss in our proposed MLR by calculating the cosine similarity between $\mathbf{\hat{y}}_{t+i}$ and $\mathbf{\bar{y}}_{t+i}$, which can be formulated below:
\begin{equation}
    \mathcal{L}_{mlr}=1-\frac{1}{K}\sum_{i=0}^{K-1}{\frac{\mathbf{\hat{y}}_{t+i}}{{\left \| \mathbf{\hat{y}}_{t+i} \right \|}_2} \frac{\mathbf{\bar{y}}_{t+i}}{{\left \| \mathbf{\bar{y}}_{t+i} \right \|}_2}}.
\end{equation}
The loss $\mathcal{L}_{mlr}$ is used to update the parameters of the online networks, including encoder $f$, predictive latent decoder $\phi$, projection head $g$ and prediction head $q$. Through our proposed self-supervised auxiliary objective \ourname, the learned state representations by the encoder will be more informative and thus can further facilitate policy learning.

\textbf{Objective.} The proposed \ourname~\tco{serves as} an auxiliary task, which is optimized together with the policy learning. Thus, the overall loss function $\mathcal{L}_{total}$ for RL agent training is:
\begin{equation}
    \mathcal{L}_{total}=\mathcal{L}_{rl}+\lambda\mathcal{L}_{mlr},
    \label{equ:total_loss}
\end{equation}
where $\mathcal{L}_{rl}$ and $\mathcal{L}_{mlr}$ are the loss functions of the base RL agent (\egno, SAC \cite{haarnoja2018soft} and Rainbow \cite{hessel2018rainbow}) and the proposed mask-based latent reconstruction, respectively. $\lambda$ is a hyperparameter for balancing the two terms. Notably, the agent of vision-based RL commonly consists of two parts, \ieno, the (state) representation network (\ieno, encoder) and the policy learning network. The encoder of \ourname~is taken as the representation network to encode observations into the state representations for RL training.
% The latent decoder \tco{can be discarded during testing since} the unmasked observations are \tco{only needed} as the inputs \tco{for training}.
The latent decoder \tco{can be discarded during testing since it is only needed for the optimization with our proposed auxiliary objective during training.}
% the unmasked observations are \tco{only needed} as the inputs \tco{for training}.
More details can be found in Appendix \ref{appendix_implementation}.
% The latent decoder is not adopted because the unmasked observations are taken as the inputs. More details can be found in Appendix \ref{appendix_implementation}.

% \vspace{-3pt}
\section{Experiment}
\subsection{Setup}
\textbf{Environments and Evaluation.} We evaluate the sample efficiency of our MLR on both the continuous control benchmark DeepMind Control Suite (DMControl) \cite{tassa2018deepmind} and the discrete control benchmark Atari \cite{bellemare2013arcade}. On DMControl, following the previous works \cite{laskin2020curl,yarats2021image,laskin2020reinforcement,yu2021playvirtual}, we choose six commonly used environments from DMControl, \ieno, \textit{Finger, spin}; \textit{Cartpole, swingup}; \textit{Reacher, easy}; \textit{Cheetah, run}; \textit{Walker, walk} and \textit{Ball in cup, catch} for the evaluation. We test the performance of RL agents with the mean and median scores over 10 episodes at 100k and 500k environment steps, referred to as \textbf{DMControl-100k} and \textbf{DMControl-500k} benchmarks, respectively.
% Concretely, DMControl-100k is used to measure sample efficiency performance, while DMControl-500k is used to measure asymptotic performance \cite{laskin2020curl}.
The score of each environment ranges from 0 to 1000 \cite{tassa2018deepmind}. 
For discrete control, we test agents on the \textbf{Atari-100k} benchmark \cite{Kaiser2020Model,van2019der,kielak2020recent,laskin2020curl} which contains 26 Atari games and allows the agents 100k interaction steps (\ieno, 400k environment steps with action repeat of 4) for training. Human-normalized score (HNS) \footnote{HNS is calculated by $\frac{S_A-S_R}{S_H-S_R}$, where $S_A$, $S_R$ and $S_H$ are the scores of the agent, random play and the expert human, respectively.} is used to measure the performance on each game. Considering the high variance of the scores on this benchmark, we test each run over 100 episodes \cite{schwarzer2021dataefficient}. To achieve a more rigorous evaluation on high-variance benchmarks such as Atari-100k with a few runs, recent work Rliable \cite{agarwal2021deep} systematically studies the evaluation bias issue in deep RL and recommends robust and efficient aggregate metrics to evaluate the overall performance (across all tasks and runs), \egno, interquartile-mean (IQM) and optimality gap (OG) \footnote{IQM discards the top and bottom 25\% of the runs and calculates the mean score of the remaining 50\% runs. OG is the amount by which the agent fails to meet a minimum score of the default human-level performance. Higher IQM and lower OG are better.}, with percentile confidence intervals (CIs, estimated by the percentile bootstrap with stratified sampling). We follow \tco{aforementioned common practices}
% the recommendation
and report the aggregate metrics on the Atari-100k benchmark with 95\% CIs.

\textbf{Implementation.} SAC \cite{haarnoja2018soft} and Rainbow \cite{hessel2018rainbow} are taken as the base continuous-control agent and discrete-control agent, respectively (see Appendix \ref{appendix_bkgd_sac} and \ref{appendix_bkgd_rainbow} for the details of the two algorithms). In our experiments, we denote the base agents trained only by RL loss $\mathcal{L}_{rl}$ (as in Equation \ref{equ:total_loss}) as \textit{Baseline}, while denoting the models of applying our proposed MLR to the base agents as \textit{MLR} for the brevity. Note that compared to naive SAC or Rainbow, our \textit{Baseline} additionally adopts data augmentation (random crop and random intensity). We adopt this following the prior works \cite{laskin2020reinforcement,yarats2021image}
%This practice follows prior works \cite{laskin2020reinforcement,yarats2021image}
which uncovers that applying proper data augmentation can significantly improve the sample efficiency of SAC or Rainbow.
As shown in Equation \ref{equ:total_loss}, we set a weight $\lambda$ to balance $\mathcal{L}_{rl}$ and $\mathcal{L}_{mlr}$ so that the gradients of these two loss items lie in a similar range and empirically find $\lambda=1$ works well in most environments. In \ourname, by default, 
% we set the length of a sampled trajectory $K$ to 16, mask ratio $\eta$ to 50\% and the size of the masked cube ($h \times w \times k$) to $10 \times 10 \times 8$ (except for $k=4$ in Cart.swin and Reac.easy due to their large motion range). 
we set the length of a sampled trajectory $K$ to 16 and mask ratio $\eta$ to 50\%. We set the size of the masked cube ($k \times h \times w$) to $8 \times 10 \times 10$ on most DMControl tasks and $8 \times 12 \times 12$ on the Atari games. 
More implementation details can be found in Appendix \ref{appendix_implementation}.
\begin{table*}[t]
    % \vspace{0pt}
    \footnotesize % control font size
    \centering
    \caption{Comparison results (mean $\pm$ std) on the DMControl-100k and DMControl-500k benchmarks. Our method augments \textit{Baseline} with the proposed \ourname~objective (denoted as \textit{\ourname}).}
    %\caption{Scores (mean and standard deviation) achieved by different methods on the DMControl after 100k environment steps (DMControl-100k) and 500k environment steps (DMControl-500k). We run our \ourname~with 10 random seeds.}
    % \caption{Results (mean $\&$ standard deviation) of different methods on the DMControl at 100k environment steps (DMControl-100k) and 500k environment steps (DMControl-500k). The result of our method is averaged over 10 random seeds.}
    \label{table:dmc_compare}
    % \vspace{-5pt}
    \scalebox{0.72}{
        \begin{tabular}{l c c c c c c c c c c}
            \toprule
            \textbf{100k Step Scores} & \textbf{PlaNet} & \textbf{Dreamer} & \textbf{SAC+AE} & \textbf{SLAC} & \textbf{CURL} & \textbf{DrQ} & \textbf{PlayVirtual} & \textbf{Baseline}  & \textbf{\ourname}\\ \hline
            \specialrule{0em}{1.5pt}{1pt}   % control space gap
            Finger, spin & 136 $\pm$ 216  & 341 $\pm$ 70   & 740 $\pm$ 64    & 693 $\pm$ 141 & 767 $\pm$ 56  & 901 $\pm$ 104  & \textbf{915 $\pm$ 49}  & 853 $\pm$ 112 & 907 $\pm$ 58 \\
            Cartpole, swingup & 297 $\pm$ 39   & 326 $\pm$ 27   & 311 $\pm$ 11    & -             & 582 $\pm$ 146 &  759 $\pm$ 92 & \textbf{816 $\pm$ 36}   & 784 $\pm$ 63 & 806 $\pm$ 48\\
            Reacher, easy & 20 $\pm$ 50    & 314 $\pm$ 155  & 274 $\pm$ 14    & -             & 538 $\pm$ 233  & 601 $\pm$ 213 & 785 $\pm$ 142   & 593 $\pm$ 118 & \textbf{866 $\pm$ 103}\\
            Cheetah, run & 138 $\pm$ 88   & 235 $\pm$ 137  & 267 $\pm$ 24    & 319 $\pm$ 56  & 299 $\pm$ 48  & 344 $\pm$ 67  & 474 $\pm$ 50   & 399 $\pm$ 80 & \textbf{482 $\pm$ 38}\\
            Walker, walk  & 224 $\pm$ 48   & 277 $\pm$ 12   & 394 $\pm$ 22    & 361 $\pm$ 73  & 403 $\pm$ 24  & 612 $\pm$ 164  & 460 $\pm$ 173   & 424 $\pm$ 281 & \textbf{643 $\pm$ 114}\\
            Ball in cup, catch & 0 $\pm$ 0      & 246 $\pm$ 174  & 391 $\pm$ 82    & 512 $\pm$ 110 & 769 $\pm$ 43  & 913 $\pm$ 53 & 926 $\pm$ 31 & 648 $\pm$ 287 & \textbf{933 $\pm$ 16}\\ 
            \midrule
            Mean & 135.8 & 289.8 & 396.2 & 471.3 & 559.7 & 688.3 & 729.3 & 616.8 & \textbf{772.8} \\ 
            Median & 137.0 & 295.5 & 351.0 & 436.5 & 560.0 & 685.5 & 800.5 & 620.5 & \textbf{836.0} \\ 
            \midrule
            \textbf{500k Step Scores}  & & & & & & & & \\ \midrule
            Finger, spin & 561 $\pm$ 284  & 796 $\pm$ 183  & 884 $\pm$ 128   & 673 $\pm$ 92  & 926 $\pm$ 45 & 938 $\pm$ 103 & 963 $\pm$ 40 & 944 $\pm$ 97 & \textbf{973 $\pm$ 31}\\
            Cartpole, swingup & 475 $\pm$ 71   & 762 $\pm$ 27   & 735 $\pm$ 63    & -             & 841 $\pm$ 45& 868 $\pm$ 10 & 865 $\pm$ 11 & 871 $\pm$ 4 & \textbf{872 $\pm$ 5} \\
            Reacher, easy & 210 $\pm$ 390  & 793 $\pm$ 164  & 627 $\pm$ 58    & -             & 929 $\pm$ 44 & 942 $\pm$ 71&942 $\pm$ 66 & 943 $\pm$ 52 & \textbf{957 $\pm$ 41}\\
            Cheetah, run & 305 $\pm$ 131  & 570 $\pm$ 253  & 550 $\pm$ 34    & 640 $\pm$ 19  & 518 $\pm$ 28  & 660 $\pm$ 96 & \textbf{719 $\pm$ 51}  & 602 $\pm$ 67 & 674 $\pm$ 37\\
            Walker, walk  & 351 $\pm$ 58   & 897 $\pm$ 49   & 847 $\pm$ 48    & 842 $\pm$ 51  & 902 $\pm$ 43  & 921 $\pm$ 4575 &928 $\pm$ 30 & 818 $\pm$ 263& \textbf{939 $\pm$ 10}\\
            Ball in cup, catch  & 460 $\pm$ 380  & 879 $\pm$ 87   & 794 $\pm$ 58    & 852 $\pm$ 71 & 959 $\pm$ 27 & 963 $\pm$ 9 & \textbf{967 $\pm$ 5} & 960 $\pm$ 10& 964 $\pm$ 14\\ 
            \midrule
            Mean & 393.7 & 782.8 & 739.5 & 751.8 & 845.8 & 882.0 & \textbf{897.3} & 856.3 & 896.5 \\ 
            Median & 405.5 & 794.5 & 764.5 & 757.5 & 914.0 & 929.5 & 935.0 & 907.0 & \textbf{948.0} \\
            \bottomrule
        \end{tabular}
    }
    % \vspace{-5pt}
\end{table*}

% \vspace{-7pt}
\subsection{Comparison with State-of-the-Arts}
\textbf{DMControl.} We compare our proposed \ourname~with the state-of-the-art (SOTA) sample-efficient RL methods proposed for continuous control, including PlaNet \cite{hafner2019learning}, Dreamer \cite{Hafner2020Dream}, SAC+AE \cite{yarats2021improving}, SLAC \cite{lee2020slac}, CURL \cite{laskin2020curl}, DrQ \cite{yarats2021image} and PlayVirtual \cite{yu2021playvirtual}. The comparison results are shown in Table \ref{table:dmc_compare}, and all results are averaged over 10 \tco{runs with different} random seeds. We can observe that: (i) \ourname~significantly improves \textit{Baseline} in both DMControl-100k and DMControl-500k, and achieves consistent gains relative to \textit{Baseline} across all environments. It is worthy to mention that, for the DMControl-100k, our proposed method outperforms \textit{Baseline} by \textbf{25.3\%} and \textbf{34.7\%} in mean and median scores, respectively. This demonstrates the superiority of \ourname~in improving the sample efficiency of RL algorithms. (ii) The RL agent equipped with \ours~outperforms most state-of-the-art methods on DMControl-100k and DMControl-500k. Specifically, our method surpasses the best previous method (\ieno, PlayVirtual) by 43.5 and 35.5 respectively on mean and median scores on DMControl-100k. Our method also delivers the best median score and reaches a comparable mean score with the strongest SOTA method on DMControl-500k.
% Note that PlayVirtual focuses on how to generate more trajectories for enhancing representation learning, while MLR focuses on how to exploit the data more efficiently. They are in fact complementary theoretically.
% Note that PlayVirtual \cite{yu2021playvirtual} generates virtual trajectories for RL training rather than designing auxiliary task to improve the representations like what we focus on in this paper. We are in fact complementary theoretically.

% \begin{wrapfigure}{r}{0.5\textwidth}
\begin{figure}[t]%[h]
    % \vspace{-10pt}
    % \vspace{-2mm}
	\begin{center}
		\includegraphics[scale=0.58]{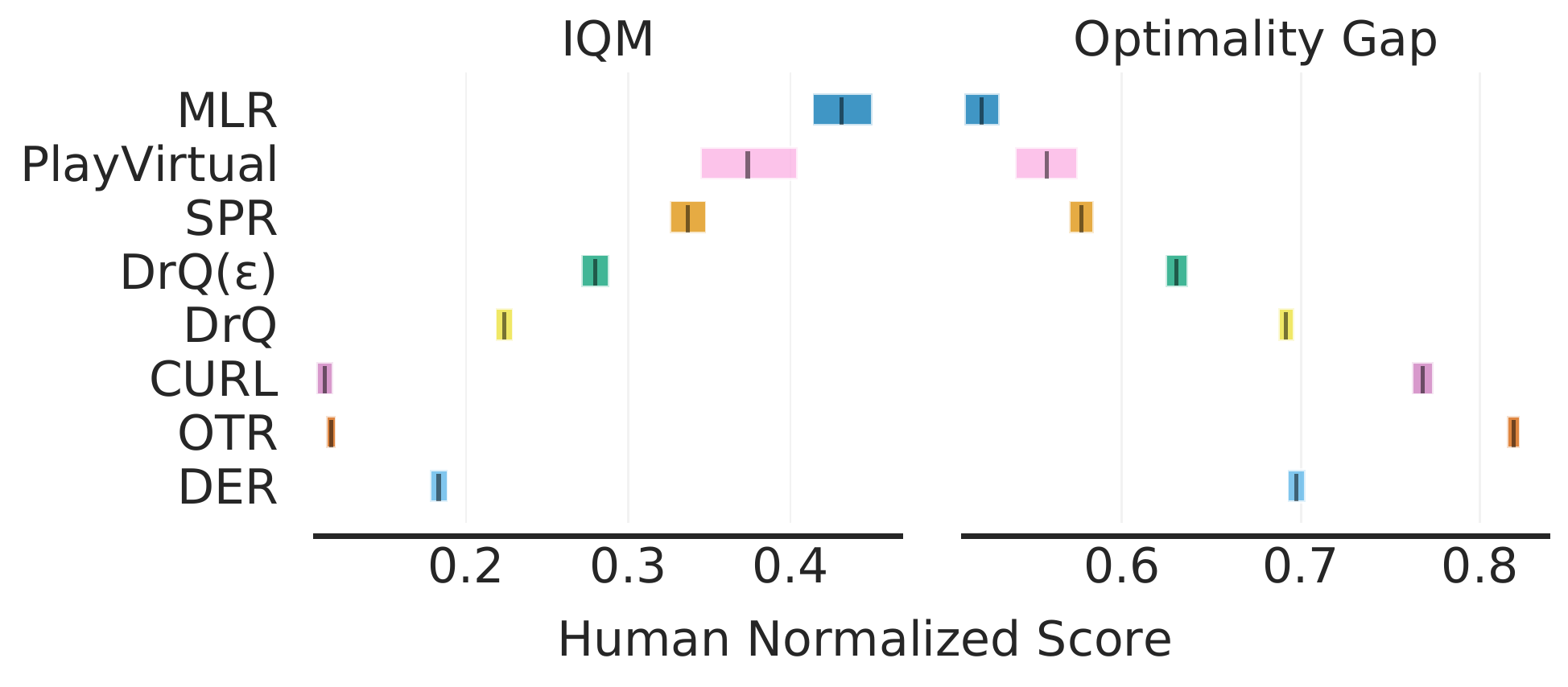} 
	\end{center}
	\vspace{-5pt}
	\caption{Comparison results on Atari-100k. Aggregate metrics (IQM and optimality gap (OG)) \cite{agarwal2021deep} with 95\% confidence intervals (CIs) are used for the evaluation. Higher IQM and lower OG are better.
	}
	\label{fig:atari-100k}
% 	\vspace{-8pt}
\end{figure}
% \end{wrapfigure}

% \begin{figure}
%      \centering
%      \begin{subfigure}[b]{0.32\textwidth}
%          \centering
%          \includegraphics[width=\textwidth]{Figs/atari_100k_profile.pdf}
%         %  \caption{$y=x$}
%         %  \label{fig:y equals x}
%      \end{subfigure}
%      \hfill
%      \begin{subfigure}[b]{0.32\textwidth}
%          \centering
%          \includegraphics[width=\textwidth]{Figs/atari_100k_profile.pdf}
%         %  \caption{$y=3sinx$}
%         %  \label{fig:three sin x}
%      \end{subfigure}
%     %  \hfill
%     %  \begin{subfigure}[b]{0.32\textwidth}
%     %      \centering
%     %      \includegraphics[width=\textwidth]{Figs/atari_100k_profile.pdf}
%     %     %  \caption{$y=3sinx$}
%     %     %  \label{fig:three sin x}
%     %  \end{subfigure}
%     %  \hfill
%     %  \begin{subfigure}[b]{0.3\textwidth}
%     %      \centering
%     %      \includegraphics[width=\textwidth]{graph3}
%     %      \caption{$y=5/x$}
%     %      \label{fig:five over x}
%     %  \end{subfigure}
%     %     \caption{Three simple graphs}
%     %     \label{fig:three graphs}
% 	\caption{Comparison results on Atari-100k. Aggregate metrics (IQM and optimality gap (OG)) \cite{agarwal2021deep} with 95\% confidence intervals (CIs) are used for the evaluation. Higher IQM and lower OG are better.
% 	}
% \end{figure}

\textbf{Atari-100k.} We further compare MLR with the SOTA model-free methods for discrete control, including DER \cite{van2019der}, OTR \cite{kielak2020recent}, CURL \cite{laskin2020curl}, DrQ \cite{yarats2021image}, DrQ($\epsilon$) (DrQ using the $\epsilon$-greedy parameters in \cite{castro2018dopamine})
% \footnote{DrQ($\epsilon$) differs from DrQ by using the $\epsilon$-greedy parameters in \cite{castro2018dopamine}.}
, SPR \cite{schwarzer2021dataefficient} and PlayVirtual \cite{yu2021playvirtual}. These methods and our MLR for Atari are all based on Rainbow \cite{hessel2018rainbow}. The Atari-100k results are shown in Figure \ref{fig:atari-100k}. 
MLR achieves an interquartile-mean (IQM) score of 0.432, which is \textbf{28.2\%} higher than SPR (IQM: 0.337) and \textbf{15.5\%} higher than PlayVirtual (IQM: 0.374). This indicates that MLR has the highest sample efficiency overall.
For the optimality gap (OG) metric, MLR reaches an OG of 0.522 \tco{which is} better than SPR (OG: 0.577) and PlayVirtual (OG: 0.558), showing that our method performs closer to the desired target, \ieno, human-level performance. 
% This shows that \ours~has superior performance on Atari.
% The full scores (with 3 random seeds) of MLR on each Atari game can be found in \tcr{Appendix}.
% The complete scores (over 3 random seeds) of MLR on the 26 Atari games and the details of the results of the previous methods can be found in \tcr{Appendix D}.
The full scores of MLR (over 3 random seeds) across the 26 Atari games and more comparisons and analysis can be found in Appendix \ref{appendix_atari}.

\subsection{Ablation Study}
In this section, we investigate the effectiveness of our MLR auxiliary objective, the impact of masking \tco{strategies}
% methods
and the model/\tcb{method} design. Unless otherwise specified, we conduct the ablation studies on the DMControl-100k benchmark and run each model \tco{with} 
% for
5 random seeds.

\textbf{Effectiveness evaluation.} Besides the comparison with SOTA methods, we demonstrate the effectiveness of our proposed \ours~by studying its improvements compared to our \textit{Baseline.} The numerical results are presented in Table \ref{table:dmc_compare}, while the curves of test performance during the training process are given in Appendix \ref{appendix_ablation}. Both numerical results and test performance curves can demonstrate that our method obviously outperforms \textit{Baseline} across different environments thanks to more informative representations learned by \ours.

\textbf{Masking strategy.} We compare three design choices of the masking operation: (i) \textit{Spatial masking} (denoted as \textit{\ourname-S}): we randomly mask \textit{patches} for each frame independently. (ii) \textit{Temporal masking} (denoted as \textit{\ourname-T}): we divide the input observation sequence into multiple segments along the temporal dimension and mask out a portion of segments randomly. (Here, the segment length is set to be equal to the temporal length of cube, \ieno, $k$.) (iii) \textit{Spatial-temporal masking} (also referred to as ``cube masking''): as aforementioned and illustrated in Figure \ref{fig:cube}, we rasterize the observation sequence into non-overlapping cubes and randomly mask a portion of them. Except for the differences described above, other configurations for masking remain the same as our proposed spatial-temporal (\ieno, cube) masking. From the results in Table \ref{table:ablate_strategy_target}, we have the following observations: (i) All three masking strategies (\ieno, \textit{\ourname-S}, \textit{\ourname-T} and \textit{\ourname}) achieve mean score improvements compared to \textit{Baseline} by \textbf{18.5\%}, \textbf{12.2\%} and \textbf{25.0\%}, respectively, and achieve median score improvements by \textbf{23.4\%}, \textbf{25.0\%} and \textbf{35.9\%}, respectively. This demonstrates the effectiveness of the core idea of introducing mask-based reconstruction to improve the representation learning of RL. (ii) Spatial-temporal masking is the most effective strategy over these three design choices. This strategy matches better with the nature of video data due to its spatial-temporal continuity in masking. It encourages the state representations to be more predictive and consistent along the spatial and temporal dimensions, thus facilitating policy learning in RL.

\textbf{Reconstruction target.} In masked language/image modeling, reconstruction/prediction is commonly performed in the original signal space, such as word embeddings or pixels. To study the influence of the reconstruction targets for the task of RL, we compare two different reconstruction spaces: (i) \textit{Pixel space reconstruction} (denoted as \textit{\ourname-Pixel}): we predict the masked contents directly by reconstructing original pixels, like the practices in CV and NLP domains; (ii) \textit{Latent space reconstruction} (\ieno, \textit{\ourname}): we reconstruct the state representations (\ieno, features) of original observations from masked observations, as we proposed in \ours. Table \ref{table:ablate_strategy_target} shows the comparison results. The reconstruction in the latent space is superior to that in the pixel space in improving sample efficiency. As discussed in the preceding sections, vision data might contain distractions and redundancies for policy learning, making the pixel-level reconstruction unnecessary. Besides, latent space reconstruction is more conducive to the coordination between the representation learning and the policy learning in RL, because the state representations are directly optimized.

\begin{table*}[t]
    \vspace{0pt}
    \footnotesize % control font size
    \centering
    \caption{Ablation study of masking strategy and reconstruction target. 
    We compare three masking strategies: spatial masking (\textit{\ourname-S}), temporal masking (\textit{\ourname-T}) and spatial-temporal masking (\textit{\ourname}), and two reconstruction targets: original pixels (denoted as \textit{\ourname-Pixel}) and momentum projections in the latent space (\ieno, \textit{\ourname}).}
    \label{table:ablate_strategy_target}
    \scalebox{1}{
        \begin{tabular}{l c c c c c}
            \toprule
            \textbf{Environment} & \textbf{Baseline} & \textbf{\ourname-S} & \textbf{\ourname-T} & \textbf{\ourname-Pixel} & \textbf{\ourname} \\ \hline
            \specialrule{0em}{1.5pt}{1pt}   % control space gap
            Finger, spin & 822 $\pm$ 146 & \textbf{919 $\pm$ 55}   & 787 $\pm$ 139   &  782 $\pm$ 95    & 907 $\pm$ 69\\
            Cartpole, swingup & 782 $\pm$ 74 & 665 $\pm$ 118     & \textbf{829 $\pm$ 33}  &  803 $\pm$ 91    & 791 $\pm$ 50\\
            Reacher, easy & 557 $\pm$ 137 & 848 $\pm$ 82    & 745 $\pm$ 84            &  787 $\pm$ 136    & \textbf{875 $\pm$ 92} \\
            Cheetah, run & 438 $\pm$ 33 & 449 $\pm$ 46      & 443 $\pm$ 43            &  346 $\pm$ 84    & \textbf{495 $\pm$ 13} \\
            Walker, walk & 414 $\pm$ 310 & 556 $\pm$ 189   & 393 $\pm$ 202           &  490 $\pm$ 216    & \textbf{597 $\pm$ 102}\\
            Ball in cup, catch & 669 $\pm$ 310 &  927 $\pm$ 6  & 934 $\pm$ 29             &  675 $\pm$ 292    & \textbf{939 $\pm$ 9} \\ 
            \midrule
            Mean & 613.7 & 727.3 & 688.5   &  647.2   & \textbf{767.3} \\ 
            Median & 613.0 & 756.5 & 766.0 &  728.5   & \textbf{833.0} \\ 
            \midrule
        \end{tabular}
    }
    % \vspace{-10pt}
\end{table*}

% \input{Tables/ablation_targets}

% \begin{wrapfigure}{r}{0.5\textwidth}
% % \begin{figure}[h]%[h] or [t]
%     \vspace{-25pt}
% 	\begin{center}
% 	\scalebox{0.7}{
% 		\includegraphics[scale=0.55]{Figs/ratio.pdf} 
% 	}
% 	\end{center}
% 	\vspace{-5mm}
% 	\caption{Ablation study of mask ratio.
% 	}
% 	\label{fig:ablate_mask_ratio}
%     \vspace{-4pt}
% % \end{figure}
% \end{wrapfigure}

% \textbf{Mask ratio.} 

\begin{wrapfigure}{r}{0.5\textwidth}
% \begin{figure}[h]%[h] or [t]
    % \vspace{-20pt}
	\begin{center}
	\scalebox{0.5}{
		\includegraphics[scale=1]{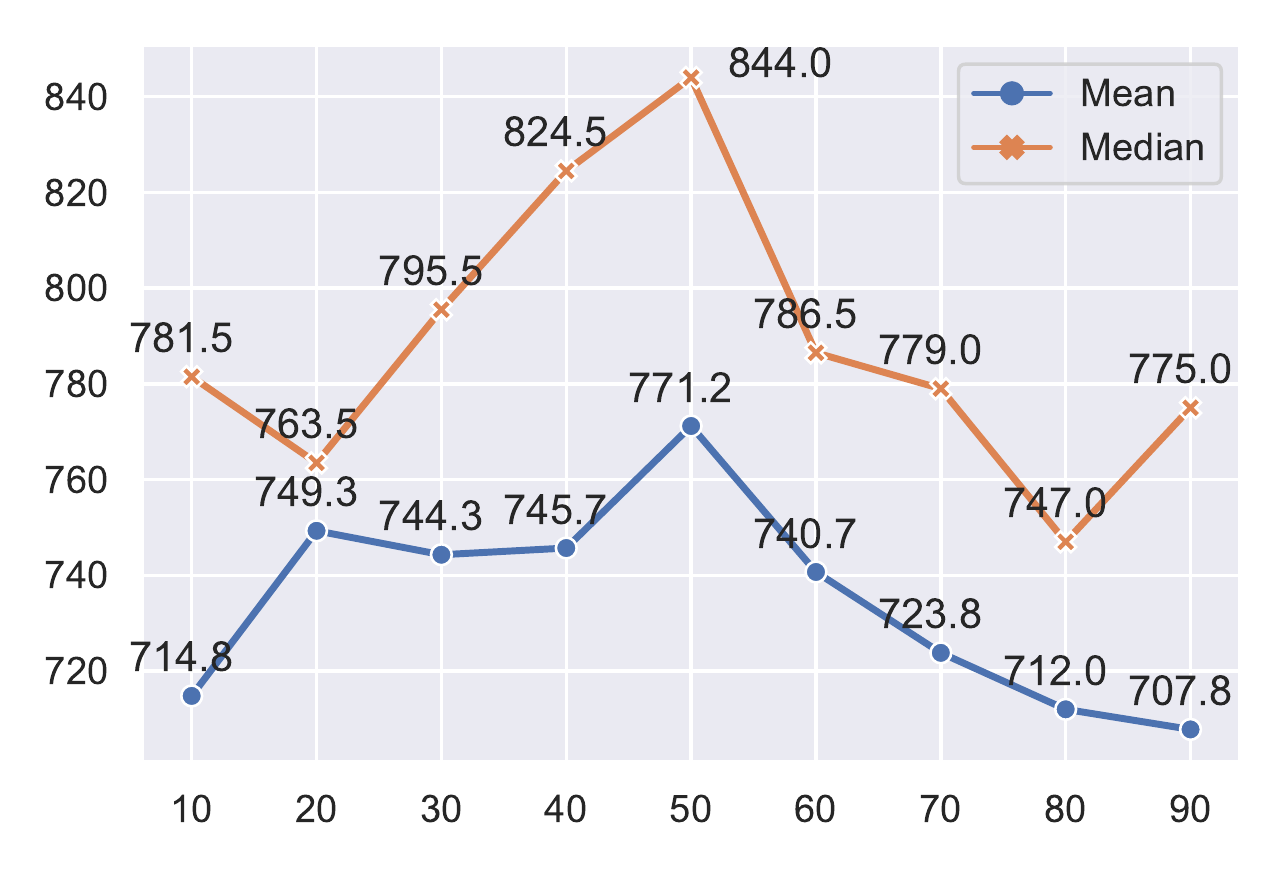} 
	}
	\end{center}
% 	\vspace{-5pt}
	\caption{Ablation study of mask ratio.
	}
	\label{fig:ablate_mask_ratio}
    % \vspace{-15pt}
% \end{figure}
\end{wrapfigure}

\textbf{Mask ratio.}
In recent works of masked image modeling \cite{he2022masked,xie2022simmim}, the mask ratio is found to be crucial for the final performance. We study the influences of different masking ratios on sample efficiency in Figure \ref{fig:ablate_mask_ratio} (with 3 random seeds) and find that the ratio of 50\% is an appropriate choice for our proposed \ours. An over-small value of this ratio could not eliminate redundancy, making the objective easy to \tco{be reached}
% reach
by extrapolation from neighboring contents that are free of capturing and understanding semantics from visual signals. An over-large value leaves few contexts for achieving the reconstruction goal. As discussed in \cite{he2022masked,xie2022simmim}, the choice of this ratio varies for different modalities and depends on the information density of the input signals.

% \textbf{Decoder depth.} We analyze the influence of using Transformer-based latent decoders of different depths. As the experimental results shown in Table \ref{table:ablate_decoder}, generally, deeper latent decoders lead to worse sample efficiency with lower mean and median scores. Notably, compared to the encoder (4.04M parameters), our decoder is lightweight (40.8K parameters). Similar to the designs in \cite{he2022masked,xie2022simmim}, it is appropriate to use a lightweight decoder in \ours, because we expect the predicting masked information to be mainly completed by the encoder instead of the decoder. Note that the state representations inferred by the encoder are adopted for the policy learning in RL. 

% \input{Tables/ablation_decoder}

\textbf{Action token.} 
We study the contributions of action tokens used for the latent decoder in Figure \ref{fig:framework} by discarding it from our proposed framework.
% (denoted as \textit{\ourname~w.o. ActTok}). 
The results are given in Table \ref{table:ablate_variants}. Intuitively, prediction only from visual signals is of more or less ambiguity. Exploiting action tokens benefits by reducing such ambiguity so that the gradients of less uncertainty can be obtained for updating the encoder.

\begin{table*}[!h]
    % \vspace{-0pt}
    \footnotesize % control font size
    \centering
    \caption{Ablation studies on action token, masking features and momentum decoder. \textit{\ourname~w.o. ActTok} denotes removing the action tokens in the input tokens of the predictive latent decoder. \textit{\ourname-F} indicates performing masking on convolutuional feature maps. \textit{\ourname-MoDec} indicates adding a momentum predictive latent decoder in the target networks.}
    \label{table:ablate_variants}
    \scalebox{1}{
        \begin{tabular}{l c c c c c}
            \toprule
            \textbf{Environment} & \textbf{Baseline} & \textbf{\ourname~w.o. ActTok} & \textbf{\ourname-F} & \textbf{\ourname-MoDec} & \textbf{\ourname}\\ \hline
            \specialrule{0em}{1.5pt}{1pt}   % control space gap
            Finger, spin & 822 $\pm$ 146 & 832 $\pm$ 46                 & 828 $\pm$ 143 & 843 $\pm$ 135 & \textbf{907 $\pm$ 69}\\
            Cartpole, swingup & 782 $\pm$ 74  & \textbf{816 $\pm$ 27}   & 789 $\pm$ 55 &  766 $\pm$ 88 & 791 $\pm$ 50\\
            Reacher, easy & 557 $\pm$ 137 & 835 $\pm$ 51                & 753 $\pm$ 159 & 800 $\pm$ 49 & \textbf{875 $\pm$ 92} \\
            Cheetah, run  & 438 $\pm$ 33  & 433 $\pm$ 72                & 477 $\pm$ 38 & 470 $\pm$ 12 & \textbf{495 $\pm$ 13} \\
            Walker, walk & 414 $\pm$ 310 & 412 $\pm$ 210                & \textbf{673 $\pm$ 33} & 571 $\pm$ 152 & 597 $\pm$ 102\\
            Ball in cup, catch & 669 $\pm$ 310 & 837 $\pm$ 114          & 843 $\pm$ 119 & 788 $\pm$ 155 & \textbf{939 $\pm$ 9} \\ 
            \midrule
            Mean & 613.7 & 694.2 & 727.2 & 706.4 & \textbf{767.3} \\ 
            Median & 613.0 & 824.2 & 771.0 & 777.0 & \textbf{833.0} \\ 
            \midrule
        \end{tabular}
    }
    % \vspace{-2mm}
\end{table*}

\textbf{Masking features.}
We compare ``masking pixels'' and ``masking features'' in Table \ref{table:ablate_variants}. Masking features (denoted by \textit{\ourname-F}) does not perform equally well compared with masking pixels as proposed in \ourname, but it still achieves significant improvements \tco{relative} to \textit{Baseline}.

\textbf{Why not use a latent decoder for targets?}
We have tried to add a momentum-updated latent decoder for obtaining the target states $\mathbf{\bar{s}}_{t+i}$ as illustrated in Figure \ref{fig:framework}. The results in Table \ref{table:ablate_variants} show that adding \tco{a} momentum decoder leads to performance drops. Since there is no information prediction for the state representation learning from the original observation sequence without masking, we do not need to refine the implicitly predicted information like that in the outputs $\mathbf{\hat{s}}_{t+i}$ of the online encoder.

\begin{wraptable}{r}{0.5\textwidth}
% \begin{table*}[t]
    % \vspace{-15pt}
    \footnotesize % control font size
    \centering
    \caption{Ablation study of predictive latent decoder depth. We report the number of parameters, mean and median scores on DMControl-100k.}
    \label{table:ablate_decoder}
    \scalebox{1}{
        \begin{tabular}{c c c c}
            \toprule
            \textbf{Layers} & \textbf{Param.} & \textbf{Mean} & \textbf{Median}\\ \hline
            \specialrule{0em}{1.5pt}{1pt}   % control space gap
            1 & 20.4K & 726.8 & 719.0\\
            2 & 40.8K & \textbf{767.3} & \textbf{833.0}\\
            4 & 81.6K & 766.2 & 789.5\\
            8 & 163.2K & 728.3 & 763.5\\
            \midrule
        \end{tabular}
    }
    % \vspace{-10pt}
% \end{table*}
\end{wraptable}
\textbf{Decoder depth.} 
% We analyze the influence of using Transformer-based latent decoders of different depths. The experimental result is shown in Appendix \ref{appendix_ablation}. Generally, deeper latent decoders lead to worse sample efficiency with lower mean and median scores. 
% Notably, compared to the encoder (4.04M parameters), our decoder is lightweight (40.8K parameters). Similar to the designs in \cite{he2022masked,xie2022simmim}, it is appropriate to use a lightweight decoder in \ours, because we expect the predicting masked information to be mainly completed by the encoder instead of the decoder. Note that the state representations inferred by the encoder are adopted for the policy learning in RL. 
We analyze the influence of using Transformer-based latent decoders with different depths. We show the experimental results in Table \ref{table:ablate_decoder}. Generally, deeper latent decoders lead to worse sample efficiency with lower mean and median scores.
% Notably, compared to the encoder (4.04M parameters), our decoder is lightweight (40.8K parameters). 
Similar to the designs in \cite{he2022masked,xie2022simmim}, it is appropriate to use a lightweight decoder in \ourname, because we expect the predicting masked information to be mainly completed by the encoder instead of the decoder. Note that the state representations inferred by the encoder are adopted for the policy learning in RL.

\textbf{Similarity loss.} We compare models using two kinds of similarity metrics to measure the distance in the latent space and observe that using cosine similarity loss is better than using mean squared error (MSE) loss. The results and analysis can be found in Appendix \ref{appendix_ablation}.

\textbf{Projection and prediction heads.} Previous works \cite{grill2020byol,schwarzer2021dataefficient,chen2020simple} have shown that in self-supervised learning, supervising the feature representations in the projected space via the projection/prediction heads is often better than in the original feature space. We investigate the effect of the two heads and find that both improve agent performance (see Appendix \ref{appendix_ablation}).

\textbf{Sequence length and cube size.} These two factors can be viewed as hyperparameters. Their corresponding experimental analysis and results are in Appendix \ref{appendix_ablation}.

\subsection{More Analysis} %Discussion}
We make more detailed investigation and analysis of our MLR from the following aspects with details shown in Appendix \ref{appendix_discussion}. (i) We provide a more detailed analysis of the effect of each masking strategy. (ii) The quality of the learned representations are further evaluated through a pretraining evaluation and a regression accuracy test. (iii) We also investigate the performance of MLR on more challenging control tasks and find that MLR is still effective. (iv) We additionally discuss the relationship between PlayVirtual and MLR. They \tco{both} achieve leading performance in \tco{improving} sample efficiency \tco{but} from different perspectives. (v) We discuss the applications and limitations of MLR. We observe that MLR is more effective on tasks with backgrounds and viewpoints that do not change drastically than on tasks with drastically changing backgrounds/viewpoints or vigorously moving objects.
% \tcb{To further investigate and understand the insights of our MLR, we have additional discussions in Appendix \ref{appendix_discussion}. We first provide a more detailed analysis of the effect of each masking strategy. The quality of the learned representations are further evaluated through a pretraining evaluation and a regression accuracy test. We also investigate the performance of MLR on more challenging control tasks and find that MLR is still effective. We additionally clarify the relationship between PlayVirtual and MLR, as they achieve leading performance in sample efficiency while from different perspectives. Moreover, we discuss the applications and limitations of MLR. We observe that MLR is more effective on tasks with backgrounds and viewpoints that do not change drastically than on tasks with drastically changing backgrounds/viewpoints or vigorously moving objects.}

\section{Conclusion}
In this work, we make the first effort to introduce the mask-based \tco{modeling}
% reconstruction
to RL for facilitating policy learning by improving the learned state representations. We propose \ourname, a simple yet effective self-supervised auxiliary objective to reconstruct the masked \tco{information} in the latent space. In this way, the learned state representations are encouraged to include richer and more informative features. 
Extensive experiments demonstrate the effectiveness of \ourname~and show that \ourname~achieves the state-of-the-art performance on both DeepMind Control and Atari benchmarks. We conduct a detailed ablation study for our proposed designs in \ours~and analyze their differences from that in NLP and CV domains.
We hope our proposed method can further inspire research for vision-based RL from the perspective of improving representation learning. Moreover, the \tco{concept of} masked latent reconstruction is also worthy of being explored and extended in CV and NLP \tco{fields}. We are looking forward to seeing more mutual promotion between different research fields.

% \begin{ack}
% This work was supported in part by the NSFC under Grant U1908209 and 62021001 and the National Key Research and Development Program of China 2018AAA0101400.
% \end{ack}

\begin{ack}
We thank all the anonymous reviewers for their valuable comments on our paper.
\end{ack}

\bibliography{neurips_2022}
\bibliographystyle{neurips_2022}
\newpage
%%%%%%%%%%%%%%%%%%%%%%%%%%%%%%%%%%%%%%%%%%%%%%%%%%%%%%%%%%%%
\section*{Checklist}

% %%% BEGIN INSTRUCTIONS %%%
% The checklist follows the references.  Please
% read the checklist guidelines carefully for information on how to answer these
% questions.  For each question, change the default \answerTODO{} to \answerYes{},
% \answerNo{}, or \answerNA{}.  You are strongly encouraged to include a {\bf
% justification to your answer}, either by referencing the appropriate section of
% your paper or providing a brief inline description.  For example:
% \begin{itemize}
%   \item Did you include the license to the code and datasets? \answerYes{See Section~\ref{gen_inst}.}
%   \item Did you include the license to the code and datasets? \answerNo{The code and the data are proprietary.}
%   \item Did you include the license to the code and datasets? \answerNA{}
% \end{itemize}
% Please do not modify the questions and only use the provided macros for your
% answers.  Note that the Checklist section does not count towards the page
% limit.  In your paper, please delete this instructions block and only keep the
% Checklist section heading above along with the questions/answers below.
% %%% END INSTRUCTIONS %%%

\begin{enumerate}

\item For all authors...
\begin{enumerate}
  \item Do the main claims made in the abstract and introduction accurately reflect the paper's contributions and scope?
    \answerYes{}
  \item Did you describe the limitations of your work?
    \answerYes{See Appendix \ref{appendix_discussion}.}
  \item Did you discuss any potential negative societal impacts of your work?
    \answerYes{See Appendix \ref{appendix_impact}.}
  \item Have you read the ethics review guidelines and ensured that your paper conforms to them?
    \answerYes{}
\end{enumerate}

\item If you are including theoretical results...
\begin{enumerate}
  \item Did you state the full set of assumptions of all theoretical results?
    \answerNA{}
        \item Did you include complete proofs of all theoretical results?
    \answerNA{}
\end{enumerate}

\item If you ran experiments...
\begin{enumerate}
  \item Did you include the code, data, and instructions needed to reproduce the main experimental results (either in the supplemental material or as a URL)?
    \answerYes{ \href{https://github.com/microsoft/Mask-based-Latent-Reconstruction}{https://github.com/microsoft/Mask-based-Latent-Reconstruction}.}
  \item Did you specify all the training details (e.g., data splits, hyperparameters, how they were chosen)?
    \answerYes{See Appendix \ref{appendix_implementation}.}
        \item Did you report error bars (e.g., with respect to the random seed after running experiments multiple times)?
    \answerYes{}
        \item Did you include the total amount of compute and the type of resources used (e.g., type of GPUs, internal cluster, or cloud provider)?
    \answerYes{See Appendix \ref{appendix_implementation}.}
\end{enumerate}

\item If you are using existing assets (e.g., code, data, models) or curating/releasing new assets...
\begin{enumerate}
  \item If your work uses existing assets, did you cite the creators?
    \answerYes{See Appendix \ref{appendix_implementation}.}
  \item Did you mention the license of the assets?
    \answerYes{See Appendix \ref{appendix_implementation}.}
  \item Did you include any new assets either in the supplemental material or as a URL?
    \answerYes{\href{https://github.com/microsoft/Mask-based-Latent-Reconstruction}{https://github.com/microsoft/Mask-based-Latent-Reconstruction}.}
  \item Did you discuss whether and how consent was obtained from people whose data you're using/curating?
    \answerYes{See Appendix \ref{appendix_implementation}.}
  \item Did you discuss whether the data you are using/curating contains personally identifiable information or offensive content?
    \answerYes{See Appendix \ref{appendix_implementation}.}
\end{enumerate}

\item If you used crowdsourcing or conducted research with human subjects...
\begin{enumerate}
  \item Did you include the full text of instructions given to participants and screenshots, if applicable?
    \answerNA{}
  \item Did you describe any potential participant risks, with links to Institutional Review Board (IRB) approvals, if applicable?
    \answerNA{}
  \item Did you include the estimated hourly wage paid to participants and the total amount spent on participant compensation?
    \answerNA{}
\end{enumerate}

\end{enumerate}

%%%%%%%%%%%%%%%%%%%%%%%%%%%%%%%%%%%%%%%%%%%%%%%%%%%%%%%%%%%%
\newpage

% \appendix

% \section{Appendix}

% Optionally include extra information (complete proofs, additional experiments and plots) in the appendix.
% This section will often be part of the supplemental material.

\appendix
%\section*{Appendix}
\section{Extended Background} \label{appendix_bkgd}
\subsection{Soft Actor-Critic} \label{appendix_bkgd_sac}
Soft Actor-Critic (SAC) \cite{haarnoja2018soft} is an off-policy actor-critic algorithm, which is based on the maximum entropy RL framework where the standard return maximization objective is augmented with an entropy maximization term \cite{ziebart2008maximum}. 
SAC has a soft Q-function $Q$ and a policy $\pi$.
% , parameterized by $\omega_Q$ and $\omega_\phi$ respectively. 
The soft Q-function is learned by minimizing the soft Bellman error:
% \begin{align}
%     \begin{aligned}
%     \mathit{J}(\mathit{Q})=\mathbb{E}_{t\sim \mathcal{D}}[\frac{1}{2}( &\mathit{Q}(\mathbf{s}_t,\mathbf{a}_t)\\&-(r_t+\gamma \mathit{V}_{\bar\theta}(\mathbf{s}_{t+1})))^2]
%     \end{aligned}
% \end{align}
\begin{equation}
    \mathit{J}(Q)=\mathbb{E}_{tr\sim \mathcal{D}}[(Q(\mathbf{s}_t,\mathbf{a}_t)-(r_t+\gamma \bar{V}(\mathbf{s}_{t+1}))^2],
\end{equation}
where $tr=(\mathbf{s}_t,\mathbf{a}_t,r_t,\mathbf{s}_{t+1})$ is a tuple with current state $\mathbf{s}_t$, action $\mathbf{a}_t$, successor $\mathbf{s}_{t+1}$ and reward $r_t$, $\mathcal{D}$ is the replay buffer and $\bar{\mathit{V}}$ is the target value function. $\bar{\mathit{V}}$ has the following expectation:
\begin{equation}
    \bar{V}(\mathbf{s}_{t})=\mathbb{E}_{\mathbf{a}_t\sim \pi}[{\bar{Q}{(\mathbf{s}_t,\mathbf{a}_t)}-\alpha \log\pi(\mathbf{a}_t|\mathbf{s}_t)}],
\end{equation}
where $\bar{Q}$ is the target Q-function whose parameters are updated by an exponential moving average of the parameters of the Q-function $Q$, and the temperature $\alpha$ is used to balance the return maximization and the entropy maximization.
The policy $\pi$ is represented by using the reparameterization trick and optimized by minimizing the following objective:
\begin{equation}
    \begin{aligned}
    J(\pi)=\mathbb{E}_{\mathbf{s}_t \sim \mathcal{D},\mathbf{\epsilon}_t \sim \mathcal{N}}[ \alpha \log \pi (f_{\pi}(\mathbf{\epsilon}_t;\mathbf{s}_t)|\mathbf{s}_t) - Q(\mathbf{s}_t, f_{\pi}(\mathbf{\epsilon}_t;\mathbf{s}_t))],
    \end{aligned}
\end{equation}

where $\mathbf{\epsilon}_t$ is the input noise vector sampled from Gaussian distribution $\mathcal{N}(0,I)$, and $f_{\pi}(\mathbf{\epsilon}_t;\mathbf{s}_t)$ denotes actions sampled stochastically from the policy $\pi$, \ieno, $f_{\pi}(\mathbf{\epsilon}_t;\mathbf{s}_t) \sim \tanh(\mu_\pi (\mathbf{s}_t)+\sigma_\pi (\mathbf{s}_t) \odot \mathbf{\epsilon}_t)$. SAC is shown to have a remarkable performance in continuous control \cite{haarnoja2018soft}.

\subsection{Deep Q-network and Rainbow} \label{appendix_bkgd_rainbow}
Deep Q-network (DQN) \cite{mnih2013playing} trains a neural network $\mathit{Q}_\theta$ with parameters $\theta$ to approximate the Q-function. DQN introduces a target Q-network $\mathit{Q}_{\theta'}$ for stable training. The target Q-network $\mathit{Q}_{\theta'}$ has the same architecture as the Q-network $\mathit{Q}_\theta$, and the parameters $\theta'$ are updated with $\theta$ every certain number of iterations. The objective of DQN is to minimize the squared error between the predictions of $\mathit{Q}_\theta$ and the target values provided by $\mathit{Q}_{\theta'}$:
\begin{equation}
    \mathit{J}(\mathit{Q_{\theta}})=(\mathit{Q}_\theta (\mathbf{s}_t,a_t)-(r_t+\gamma \max\limits_a \mathit{Q}_{\theta'} (\mathbf{s}_{t+1},a)))^2
\end{equation}
Rainbow \cite{hessel2018rainbow} integrates a number of improvements on the basis of the vanilla DQN \cite{mnih2013playing} including: (i) employing modified target Q-value sampling in Double DQN \cite{van2016deep}; (ii) adopting Prioritized Experience Replay \cite{schaul2015prioritized} strategy; (iii) decoupling the value function of state and the advantage function of action from Q-function like Dueling DQN \cite{wang2016dueling}; (iv) introducing distributional RL and predicting value distribution as C51 \cite{bellemare2017distributional}; (v) adding parametric noise into the network parameters like NoisyNet \cite{fortunato2017noisy}; and (vi) using multi-step return \cite{sutton2018reinforcement}. Rainbow is typically
regarded as a strong model-free baseline for discrete control.
% Further, to improve the sample efficiency of Rainbow, van Hasselt \etal \cite{van2019der} propose Data-Efficient Rainbow and Kielak \etal \cite{kielak2020recent} propose Overtrained Rainbow.

\begin{figure*}[t]%[h] or [t]
	\begin{center}
	\scalebox{0.7}{
		\includegraphics[scale=0.65]{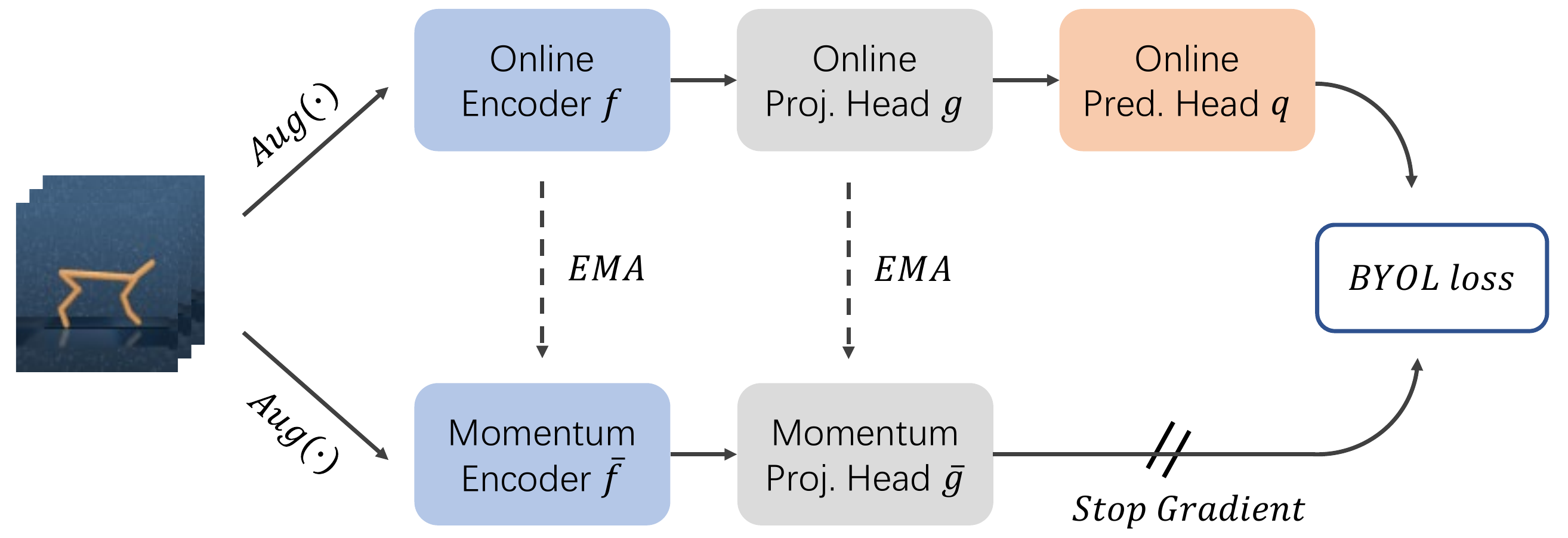} 
	}
	\end{center}
% 	\vspace{-5mm}
	\caption{An illustration of the framework of BYOL \cite{grill2020byol}.
	}
	\label{fig:byol}
% 	\vspace{-2mm}
\end{figure*}

\tcr{
\subsection{BYOL-style Auxiliary Objective} \label{appendix_byol}
BYOL \cite{grill2020byol} is a strong self-supervised representation learning method by enforcing the similarity of the representations of the same image across diverse data augmentation. The pipeline is shown in Figure \ref{fig:byol}. BYOL has an online branch and a momentum branch. The momentum branch is used for computing a stable target for learning representations \cite{he2020momentum,tarvainen2017mean}. BYOL is composed of an online encoder $f$, a momentum encoder $\bar{f}$, an online projection head $g$, a momentum projection head $\bar{g}$ and an prediction head $q$. The momentum encoder and projection head have the same architectures as the corresponding online networks and are updated by an exponential moving average (EMA) of the online weights (see Equation \ref{equ:ema} in the main manuscript). The prediction head is only used in the online branch, making BYOL's architecture asymmetric. Given an image $x$, BYOL first produces two views $v$ and $v'$ from $x$ through images augmentations. The online branch outputs a representation $y=f(v)$ and a projection $z=g(y)$, and the momentum branch outputs $y'=\bar{f}(v')$ and a target projection $z'=\bar{g}(y')$. BYOL then use a prediction head $q$ to regress $z'$ from $z$, \ieno, $q(z) \rightarrow z'$. BYOL minimizes the similarity loss between $q(z)$ and a stop-gradient\footnote{Stop gradient operation stops the gradients from passing through, avoiding trivial solutions \cite{he2020momentum}.} target $sg(z')$.
\begin{equation}
    \mathcal{L}_{BYOL}=\left \| q(z)-sg(z') \right \|^2_2 =2-2{\frac{q(z)}{{\left \| q(z) \right \|}_2} \frac{sg(z')}{{\left \| sg(z') \right \|}_2}}.
\end{equation}
Inspired by the success of BYOL in learning visual representations, recent works introduce BYOL-style learning objectives to vision-based RL for learning effective state representations and show promising performance \cite{schwarzer2021dataefficient,yu2021playvirtual,yarats2021reinforcement,hansen2021generalization,guo2022byol}. The BYOL-style learning is often integrated into auxiliary objectives in RL such as future state prediction \cite{schwarzer2021dataefficient,guo2022byol}, cycle-consistent dynamics prediction \cite{yu2021playvirtual}, prototypical representation learning \cite{yarats2021reinforcement} and invariant representation learning \cite{hansen2021generalization}. These works also show that it is more effective to supervise/regularize the predicted representations in the BYOL's projected latent space than in the representation or pixel space.
Besides, the BYOL-style auxiliary objectives are commonly trained with data augmentation since it can conveniently produce two BYOL views. For example, SPR \cite{schwarzer2021dataefficient} and PlayVirtual \cite{yu2021playvirtual} apply random crop and random intensity to input observations in Atari games. The proposed MLR auxiliary objective can be categorized into BYOL-style auxiliary objectives.
% In the practice of original BYOL and most BYOL-style auxiliary objectives in RL, the online and momentum branches share the same data augmentation transform function. Recent SODA
% SPR \cite{schwarzer2021dataefficient} introduces BYOL-style self-supervision by enforcing the predicted future states to be similar to the groundtruth states
% performs multi-step dynamics prediction in the latent space.
}

% \tcr{
% \textit{Discussion}:(i)
% }

\section{Implementation Detail} \label{appendix_implementation}
\subsection{Network Architecture} \label{appendix_arch}

Our model has two parts: the basic networks and the auxiliary networks. The basic networks are composed of a representation network (\ieno, encoder) $f$ parameterized by $\theta_f$ and the policy learning networks $\omega$ (\egno, SAC \cite{haarnoja2018soft} or Rainbow \cite{hessel2018rainbow})  parameterized by $\theta_\omega$. 

We follow CURL \cite{laskin2020curl} to build the architecture of the basic networks on the DMControl \cite{tassa2018deepmind} benchmarks. The encoder is composed of four convolutional layers (with a rectified linear units (ReLU) activation after each), a fully connected (FC) layer, and a layer normalization (LN) \cite{ba2016layer} layer. Furthermore, the policy learning networks are built by multilayer perceptrons (MLPs). For the basic networks on Atari \cite{bellemare2013arcade}, we also follow CURL to adopt the original architecture of Rainbow \cite{hessel2018rainbow} where the encoder consists of three convolutional layers (with a ReLU activation after each), and the Q-learning heads are MLPs.

Our auxiliary networks have online networks and momentum (or target) networks. The online networks consist of an encoder $f$, a predictive latent decoder (PLD) $\phi$, a projection head $g$ and a prediction head $q$, parameterized by $\theta_f$, $\theta_\phi$, $\theta_g$ and $\theta_q$, respectively. Notably, the encoders in the basic networks and the auxiliary networks are \textit{shared}. As shown in Figure \ref{fig:framework} in our main manuscript, there are a momentum encoder $\bar{f}$ and a momentum projection head $\bar{g}$ for computing the self-supervised targets. The momentum networks have the same architectures as the corresponding online networks. Our PLD is a transformer encoder \cite{vaswani2017attention} and has two standard attention layers (with a single attention head). We use an FC layer as the action embedding head to transform the original action into an embedding which has the same dimension as the state representation (\ieno, state token). \tcr{The transformer encoder treats each input token as independent of the other. Thus, positional embeddings, which encode the positional information of tokens in a sequence, are added to the input tokens of PLD to maintain their relative temporal positional information (\ieno, the order of the state/action sequences).} We use sine and cosine functions to build the positional embeddings following \cite{vaswani2017attention}:
\begin{equation}
    p_{(pos, 2j)} = \sin(pos/10000^{2j/d}),
\end{equation}
\vspace{-10pt}
\begin{equation}
    p_{(pos, 2j+1)} = \cos(pos/10000^{2j/d}),
\end{equation}
where $pos$ is the position, $j$ is the dimension and $d$ is the embedding size (equal to the state representation size). 
We follow PlayVirtual \cite{yu2021playvirtual} in the architecture design of the projection and the prediction heads, built by MLPs and FC layers.
% On DMControl, both the projection head and the prediction head have two FC layers with a hidden size of 100, and a ReLU activation is followed by the first FC layer. 
% We further present the parameters of the main networks in our method in Table \ref{table:params}. The encoder dominates the number of parameters in the auxiliary networks, which implies that the encoder plays a major role in the masked prediction.

\subsection{Training Detail} \label{appendix_training}
\textbf{Optimization and training.} 
The training algorithm of our method is presented in Algorithm \ref{algo: mlr}. We use Adam optimizer \cite{kingma2014adam} to optimize all trainable parameters in our model, with $(\beta_1, \beta_2)=(0.9,0.999)$ (except for $(0.5, 0.999)$ for SAC temperature $\alpha$). Modest data augmentation such as crop/shift is shown to be effective for improving RL agent performance in vision-based RL \cite{yarats2021image, laskin2020reinforcement, schwarzer2021dataefficient, yu2021playvirtual}. Following \cite{yu2021playvirtual,schwarzer2021dataefficient,laskin2020curl}, we use random crop and random intensity in training the auxiliary objective, \ieno, $\mathcal{L}_{mlr}$. Besides, we warmup the learning rate of our \ourname~objective on DMControl by
\begin{equation}
    lr=lr_0 \cdot \min({step\_num}^{-0.5}, {step\_num}\cdot{warmup\_step}^{-1.5}),
\end{equation}
where the $lr$ and $lr_0$ denote the current learning rate and the initial learning rate, respectively, and $step\_num$ and $warmup\_step$ denote the current step and the warmup step, respectively. We empirically find that the warmup schedule bring improvements on DMControl.

\renewcommand{\algorithmicrequire}{\textbf{Require:}}  % Use Input in the format of Algorithm
\begin{algorithm}[t]
%   \caption{Cycle-consistent Predictive Representations}
  \caption{Training Algorithm for \ourname}
  \label{algo: mlr}
  \begin{algorithmic}[1]
    \Require
      An online encoder $f$, a momentum encoder $\bar{f}$, a predictive latent decoder $\phi$, an online projection head $g$, a momentum projection head $\bar{g}$, a prediction head $q$ and policy learning networks $\omega$, parameterized by $\theta_f$, $\bar{\theta}_f$, $\theta_\phi$, $\theta_g$, $\bar{\theta}_g$, $\theta_q$ and $\theta_\omega$, respectively; a stochastic cube masking function $Mask(\cdot)$; a stochastic image augmentation function $Aug(\cdot)$; an optimizer $Optimize(\cdot, \cdot)$.
     \State Determine auxiliary loss weight $\lambda$, sequence length $K$, mask ratio $\eta$, cube size $k \times h \times w$ and EMA coefficient $m$.
     \State Initialize a replay buffer $\mathcal{D}$.
     \State Initialize $Mask(\cdot)$ with $\eta$ and $k \times h \times w$.
     \State Initialize all network parameters.
    %   \State let $\theta = \theta_f \cup \theta_g$, $\bar{\theta} = \bar{\theta}_f \cup \bar{\theta}_g$; $\bar{\theta}\leftarrow \theta$
      \While {$train$}
        \State Interact with the environment based on the policy
        \State Collect the transition:  $\mathcal{D}\leftarrow \mathcal{D}\cup (\mathbf{o},\mathbf{a},\mathbf{o}_{next},r)$
        \State Sample a trajectory of $K$ timesteps $\{\mathbf{o}_t,\mathbf{a}_t,\mathbf{o}_{t+1},\mathbf{a}_{t+1},\cdots, \mathbf{o}_{t+K-1},\mathbf{a}_{t+K-1}\}$ from $\mathcal{D}$
        \State Initialize losses: $\mathcal{L}_{mlr} \leftarrow 0$; $\mathcal{L}_{rl} \leftarrow 0$
        \State Randomly mask the observation sequence:
        \Statex \quad \quad \quad $\{ \mathbf{\tilde{o}}_t,\mathbf{\tilde{o}}_{t+1},\cdots, \mathbf{\tilde{o}}_{t+K-1}\} \leftarrow Mask(\{\mathbf{o}_t,\mathbf{o}_{t+1},\cdots, \mathbf{o}_{t+K-1}\})$
        \State Perform augmentation and encoding: 
        \Statex \quad \quad \quad $\{\mathbf{\tilde{s}}_t,\mathbf{\tilde{s}}_{t+1},\cdots, \mathbf{\tilde{s}}_{t+K-1}\} \leftarrow \{ f(Aug(\mathbf{\tilde{o}}_t)),f(Aug(\mathbf{\tilde{o}}_{t+1})),\cdots, f(Aug({\mathbf{\tilde{o}}_{t+K-1}})\}$
        \State Perform decoding: 
        \Statex \quad \quad \quad  $\{\mathbf{\hat{s}}_t,\mathbf{\hat{s}}_{t+1},\cdots, \mathbf{\hat{s}}_{t+K-1}\} \leftarrow \phi(\{\mathbf{\tilde{s}}_t,\mathbf{\tilde{s}}_{t+1},\cdots, \mathbf{\tilde{s}}_{t+K-1}\}; \{\mathbf{a}_t,\mathbf{a}_{t+1},\cdots,\mathbf{a}_{t+K-1}\})$
        \State Perform projection and prediction: 
        \Statex \quad \quad \quad $ \{\mathbf{\hat{y}}_t,\mathbf{\hat{y}}_{t+1},\cdots,\mathbf{\hat{y}}_{t+K-1}\} \leftarrow \{q(g(\mathbf{\hat{s}}_t)),q(g(\mathbf{\hat{s}}_{t+1})),\cdots, q(g(\mathbf{\hat{s}}_{t+K-1}))\}$
        \State Calculate targets:
        \Statex \quad \quad \quad $\{\mathbf{\bar{y}}_t,\mathbf{\bar{y}}_{t+1},\cdots,\mathbf{\bar{y}}_{t+K-1}\} \leftarrow \{\bar{g}(\bar{f}(Aug(\mathbf{o}_t))),\bar{g}(\bar{f}(Aug(\mathbf{o}_{t+1}))),$
        \Statex \quad \quad \quad \quad \quad \quad \quad \quad \quad \quad \quad \quad \quad \quad \quad ~ $\cdots, \bar{g}(\bar{f}(Aug(\mathbf{o}_{t+K-1})))\}$
        \State Calculate \ourname~loss: $\mathcal{L}_{mlr} \leftarrow  1-\frac{1}{K}\sum_{i=0}^{K-1}{\frac{\mathbf{\hat{y}}_{t+i}}{{\left \| \mathbf{\hat{y}}_{t+i} \right \|}_2} \frac{\mathbf{\bar{y}}_{t+i}}{{\left \| \mathbf{\bar{y}}_{t+i} \right \|}_2}}$
        \State Calculate RL loss $\mathcal{L}_{rl}$ based on a given base RL algorithm (\egno, SAC)
        \State Calculate total loss: $\mathcal{L}_{total} \leftarrow \mathcal{L}_{rl} + \lambda \mathcal{L}_{mlr}$
        \State Update online parameters: $(\theta_f, \theta_\phi, \theta_g, \theta_q, \theta_\omega) \leftarrow Optimize((\theta_f, \theta_\phi, \theta_g, \theta_q, \theta_\omega),\mathcal{L}_{total})$
        \State Update momentum parameters: $(\bar{\theta}_f, \bar{\theta}_g) \leftarrow m(\bar{\theta}_f, \bar{\theta}_g)+(1-m)(\theta_f,\theta_g)$
    \EndWhile
\end{algorithmic}
\end{algorithm}

% Refer to CURL Appendix!!!
\textbf{Hyperparameters.} We present all hyperparameters used for the DMControl benchmarks \cite{tassa2018deepmind} in Table \ref{table:dmc_hyperparam} and the Atari-100k benchmark in Table \ref{table:atari_hyperparam}. 
We follow prior work \cite{yu2021playvirtual,schwarzer2021dataefficient,laskin2020curl} for the policy learning hyperparameters (\ieno, SAC and Rainbow hyperparameters).
The hyperparameters specific to our MLR auxiliary objective, including \ourname~loss weight $\lambda$, mask ratio $\eta$, the length of the sampled sequence $K$, cube shape $k \times h \times w$ and the depth of the decoder $L$, are shown in the bottom of the tables. By default, we set $\lambda$ to 1, $\eta$ to 50$\%$, $L$ to 2, $k$ to 8, and $h \times w$ to $10 \times 10$ on DMControl and $12 \times 12$ on Atari. We exceptionally set $k$ to 4 in \textit{Cartpole-swingup} and \textit{Reacher-easy} on DMControl due to their large motion range, and $\lambda$ to 5 in \textit{Pong} and \textit{Up N Down} on Atari as their MLR losses are relatively smaller than the rest 24 Atari games.
% Our MLR is flexible, allowing us to extend it to the discrete control benchmark Atari with minimum modification.

% We empirically find that the hyperparameters work well in most environments.
% % Most important hyperparameters used for DMControl and Atari are consistent.
% Exceptionally, we set , and we increase $\lambda$ to 5 in \textit{Pong} and \textit{Up N Down} in Atari as their MLR loss is relatively smaller than the rest Atari games.

% \textit{Reacher-easy} due to their large motion range)

\tcr{\textbf{Baseline and data augmentation.}
% Recent advances show that modest image augmentation such as crop can effectively improve RL agent performance in vision-based RL \cite{yarats2021image, laskin2020reinforcement, schwarzer2021dataefficient, yu2021playvirtual}. 
Our baseline models (\textit{Baseline}) are equivalent to CURL \cite{laskin2020curl} without the auxiliary loss except for the slight differences on the applied data augmentation strategies. CURL adopts random crop for data augmentation. We follow the prior work SPR \cite{schwarzer2021dataefficient} and PlayVirtual \cite{yu2021playvirtual} to adopt random crop and random intensity for both our \textit{Baseline} and \ourname.
% , except that we use random shift and intensity as augmentation on Atari following \cite{schwarzer2021dataefficient,yu2021playvirtual} while CURL uses random crop.
% When introducing an auxiliary objective to \textit{Baseline}, the two objectives are trained in a decoupled way in practice \cite{laskin2020curl,yu2021playvirtual}. We train our MLR auxiliary objective with random crop (DMControl) / shift (Atari) and intensity following PlayVirtual \cite{yu2021playvirtual}. Note that crop is commonly used on DMControl while shift on Atari \cite{schwarzer2021dataefficient,yu2021playvirtual}.
% we can use different augmentations for learning the auxiliary objectives and Baseline's RL objectives 
% in the practice of training SAC on DMControl \cite{laskin2020curl,yarats2019improving,yu2021playvirtual}, the
% we perform random crop on the observations in training the RL objective $\mathcal{L}_{rl}$ and random crop/shift and intensity in training the auxiliary objective, \ieno, $\mathcal{L}_{mlr}$.
% Note that the only difference between \textit{Baseline} and \ourname~is that \ourname~augments \textit{Baseline} with the proposed auxiliary objective.
}

\textbf{GPU and wall-clock time.} In our experiment, we run MLR with a single GPU ( NVIDIA Tesla V100 or GeForce RTX 3090) for each environment. \tcr{MLR has the same inference time complexity as \textit{Baseline} since both use only the encoder and policy learning head during testing. On Atari, the average wall-clock training time is 6.0, 10.9, 4.0, and 8.2 hours for SPR, PlayVirtual, Baseline and MLR, respectively. On DMControl, the training time is 4.3, 5.2, 3.8, 6.5 hours for SPR, PlayVirtual, Baseline and MLR, respectively. We will leave the optimization of our training strategy as future work to speed up the training, \egno, reducing the frequency of using masked-latent reconstruction.}
% On the DMControl benchmarks, the average wall-clock training time per 100k environment steps for MLR is 6.5 hours. For the games on Atari-100k, the average wall-clock training time is 8.2 hours.  
% % \tcr{While adopting our MLR auxiliary objective to  significantly improve Compared to \textit{Baseline}'s 3.8 hours on DMControl and 4.0 hours on Atari.}
% \tcr{MLR increases \textit{Baseline}'s wall-clock training time (Atari: 4.0 hours; DMControl: 3.8 hours) by 4.2 hours on Atari and 2.7 hours on DMControl respectively. While the increased time is basically acceptable in practice, we also consider parallel training of MLR to reduce the training time in future work. Compared with the state-of-the-art methods such as SPR (Atari: 6.0 hours; DMControl: 4.3 hours) and PlayVirtual (Atari: 10.9 hours; DMControl: 5.2 hours), MLR bears a similar time complexity.}
The evaluation is based on a single GeForce RTX 3090 GPU.

\subsection{Environment and Code} \label{appendix_env}
DMControl \cite{tassa2018deepmind} and Atari \cite{bellemare2013arcade} are widely used environment suites in RL community, which are public and do not involve personally identifiable information or offensive contents. We use the two environment suites to evaluate model performance. The implementation of MLR is based on the open-source PlayVirtual \cite{yu2021playvirtual} codebase \footnote{Link: https://github.com/microsoft/Playvirtual, licensed under the MIT License.}. The statistical tools on Atari are obtained from the open-source library \textit{rliable} \footnote{Link: https://github.com/google-research/rliable, licensed under the Apache License 2.0.}\cite{agarwal2021deep}.

\section{More Experimental Results and Analysis} \label{appendix_exp}
\subsection{More Atari-100k Results} \label{appendix_atari}

\begin{figure*}[t]%[h] or [t]
	\begin{center}
	\scalebox{0.6}{
		\includegraphics[scale=0.65]{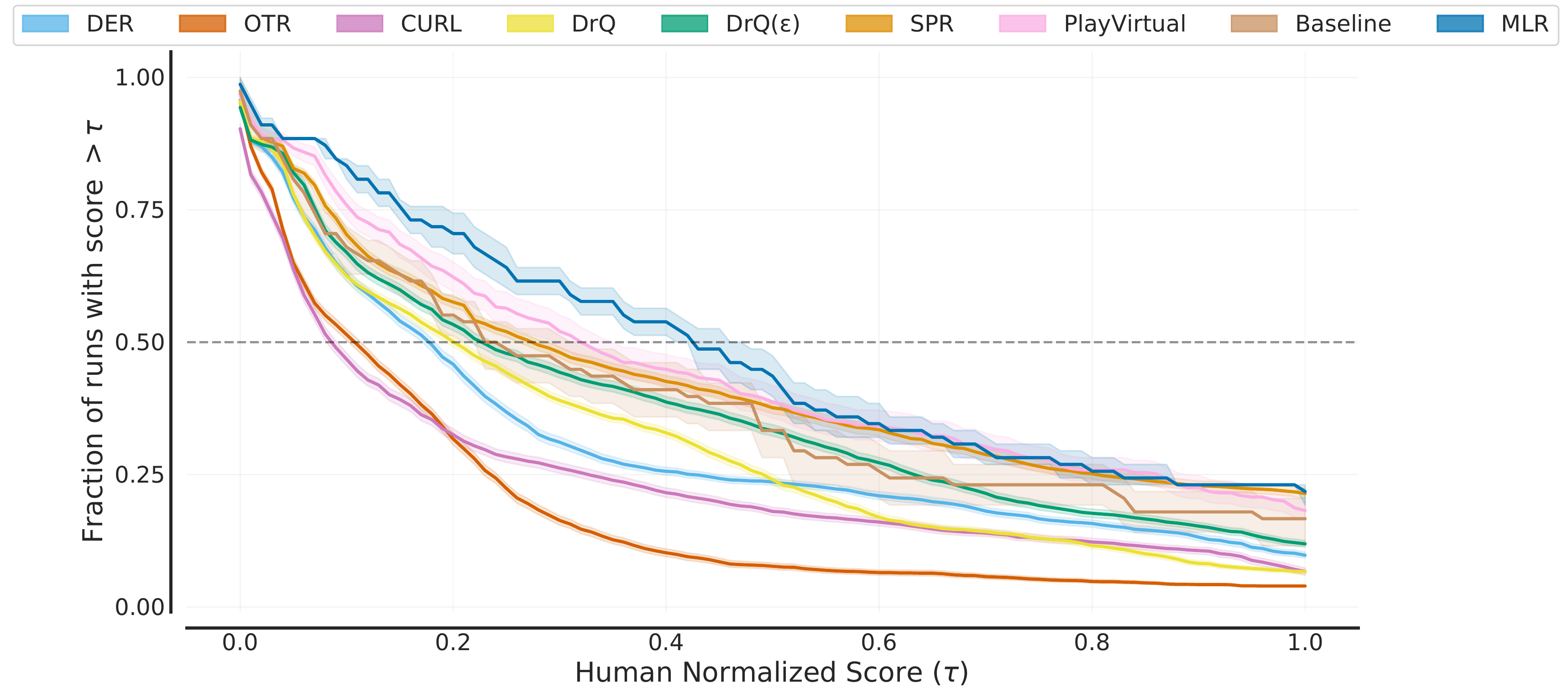} 
	}
	\end{center}
% 	\vspace{-5mm}
	\caption{Performance profiles on the Atari-100k benchmark based on human-normalized score distributions. Shaded regions indicates 95\% confidence bands. The score distribution of MLR is clearly superior to previous methods and \textit{Baseline}.
	}
	\label{fig:atari_profile}
% 	\vspace{-2mm}
\end{figure*}

We present the comparison results across all 26 games on the Atari-100k benchmark in Table \ref{table:atari_compare}. Our MLR reaches the highest scores on 11 out of 26 games and outperforms the compared methods on the aggregate metrics, \ieno, interquartile-mean (IQM) and optimality gap (OG) with 95\% confidence intervals (CIs). Notably, MLR improve the \textit{Baseline} performance by 47.9\% on IQM, which shows the effectiveness of our proposed auxiliary objective. We also present the \textit{performance profiles}\footnote{Performance profiles \cite{dolan2002benchmarking} show the tail distribution of scores on combined runs across tasks \cite{agarwal2021deep}. Performance profiles of a distribution $X$ is calculated by $\hat{F}_{X}(\tau)=\frac{1}{M} \sum_{m=1}^{M} \frac{1}{N} \sum_{n=1}^{N} \mathds{1} \left[x_{m, n}>\tau\right]$, indicating the fraction of runs above a score $\tau$ across $N$ tasks and $M$ seeds.} using human-normalized scores (HNS) with 95\% CIs in Figure \ref{fig:atari_profile}. 
The performance profiles confirm the superiority and effectiveness of our MLR. 
The results of DER \cite{van2019der}, OTR \cite{kielak2020recent}, CURL \cite{laskin2020curl}, DrQ \cite{yarats2021image} and SPR \cite{schwarzer2021dataefficient} are from \textit{rliable} \cite{agarwal2021deep}, based on 100 random seeds.
The results of PlayVirtual \cite{yu2021playvirtual} are based on 15 random seeds, and the results of \textit{Baseline} and MLR are averaged over 3 random seeds and each run is evaluated with 100 episodes. We report the standard deviations across runs of \textit{Baseline} and MLR in Table \ref{table:atari_std}.

\begin{table*}[h]
    \footnotesize % control font size
    % \scriptsize
    \centering
    %Scores achieved by different methods on the 26 games of Atari \cite{bellemare2013arcade} after 100k interaction steps ($i.e.$ Atari-100k).
    \caption{Comparison on the Atari-100k benchmark. Our method reaches the highest scores on 11 out of 26 games and the best performance concerning the aggregate metrics, \ieno, interquartile-mean (IQM) and optimality gap (OG) with 95\% confidence intervals. Our method augments \textit{Baseline} with the MLR objective and achieves a 47.9\% relative improvement on IQM.} 
    % \caption{Results of different methods on the 26 games of Atari 2600 Games \cite{bellemare2013arcade} at 100k interaction steps ($i.e.$ Atari-100k). The median human-normalized score (HNS) over the 26 games is used to measure the overall performance. Human and Random refer to expert human play and random play respectively \cite{wang2016dueling}. The result of our method result is averaged over 15 random seeds.}
    \label{table:atari_compare}
    % \vskip 0.05in    % control the gap between caption and table
    % \setlength{\tabcolsep}{2.0mm}{
    % \resizebox{\textwidth}{!}{
    \scalebox{0.77}{
        \begin{tabular}{l c c c c c c c c c c}
            \toprule
            \textbf{Game} & \textbf{Human} & \textbf{Random} & \textbf{DER} & \textbf{OTR} & \textbf{CURL} & \textbf{DrQ} & \textbf{SPR} & \textbf{PlayVirtual} & \textbf{Baseline} & \textbf{\ourname}\\
            % \midrule
            % Distance Based Models
            \midrule
            Alien           & 7127.7    & 227.8     & 802.3     & 570.8      & 711.0     & 734.1      & 841.9    & 947.8    & 678.5  & \textbf{990.1} \\
            Amidar          & 1719.5    & 5.8       & 125.9     & 77.7      & 113.7     & 94.2      & 179.7      & 165.3    & 132.8  & \textbf{227.7} \\
            Assault         & 742.0     & 222.4     & 561.5     & 330.9      & 500.9      & 479.5      & 565.6   & \textbf{702.3}    & 493.3  & 643.7 \\
            Asterix         & 8503.3    & 210.0     & 535.4    & 334.7      & 567.2     & 535.6      & 962.5     & 933.3    & \textbf{1021.3} & 883.7 \\
            Bank Heist      & 753.1     & 14.2      & 185.5      & 55.0       & 65.3      & 153.4      & \textbf{345.4}   & 245.9    & 288.2  & 180.3 \\
            Battle Zone     & 37187.5   & 2360.0    & 8977.0    & 5139.4    & 8997.8     & 10563.6    & 14834.1  & 13260.0  & 13076.7& \textbf{16080.0} \\
            Boxing          & 12.1      & 0.1       & -0.3       & 1.6        & 0.9        & 6.6        & 35.7   & \textbf{38.3}     & 14.3  & 26.4 \\
            Breakout        & 30.5      & 1.7       & 9.2      & 8.1        & 2.6      & 15.4        & 19.6      & \textbf{20.6}     & 16.7  & 16.8 \\
            Chopper Cmd & 7387.8    & 811.0         & 925.9    & 813.3      & 783.5    & 792.4     & \textbf{946.3}       & 922.4    & 878.7 & 910.7 \\
            Crazy Climber   & 35829.4   & 10780.5   & 34508.6    & 14999.3    & 9154.4    & 21991.6    & \textbf{36700.5} & 23176.7  & 28235.7 & 24633.3 \\
            Demon Attack    & 1971.0    & 152.1     & 627.6      & 681.6      & 646.5      & \textbf{1142.4}      & 517.6 & 1131.7   & 310.5  & 854.6 \\
            Freeway         & 29.6      & 0.0       & 20.9      & 11.5       & 28.3       & 17.8       & 19.3    & 16.1     & \textbf{30.9}  & 30.2 \\
            Frostbite       & 4334.7    & 65.2      & 871.0      & 224.9      & 1226.5      & 508.1     & 1170.7 & 1984.7   & 994.3  &\textbf{2381.1}  \\
            Gopher          & 2412.5    & 257.6     & 467.0      & 539.4      & 400.9      & 618.0      & 660.6  & 684.3    & 650.9  & \textbf{822.3} \\
            Hero            & 30826.4   & 1027.0    & 6226.0     & 5956.5     & 4987.7     & 3722.6     & 5858.6 & \textbf{8597.5}   & 4661.2  & 7919.3 \\
            Jamesbond       & 302.8     & 29.0      & 275.7      & 88.0      & 331.0      & 251.8      & 366.5   & 394.7    & 270.0  & \textbf{423.2} \\
            Kangaroo        & 3035.0    & 52.0      & 581.7      & 348.5      & 740.2      & 974.5      & 3617.4 & 2384.7   & 5036.0  & \textbf{8516.0} \\
            Krull           & 2665.5    & 1598.0    & 3256.9     & 3655.9     & 3049.2     & \textbf{4131.4}     & 3681.6 & 3880.7   & 3571.3  & 3923.1 \\
            Kung Fu Master  & 22736.3   & 258.5     & 6580.1    & 6659.6    & 8155.6     & 7154.5    & \textbf{14783.2}   & 14259.0  & 10517.3  & 10652.0 \\
            Ms Pacman       & 6951.6    & 307.3     & 1187.4     & 908.0     & 1064.0      & 1002.9     & 1318.4 & 1335.4   & 1320.9  & \textbf{1481.3} \\
            Pong            & 14.6      & -20.7     & -9.7       & -2.5      & -18.5        & -14.3      & -5.4  & -3.0     & -3.1  & \textbf{4.9} \\
            Private Eye     & 69571.3   & 24.9      & 72.8       & 59.6       & 81.9      & 24.8      & 86.0     & 93.9     & 93.3  & \textbf{100.0} \\
            Qbert           & 13455.0   & 163.9     & 1773.5     & 552.5     & 727.0      & 934.2     & 866.3    & \textbf{3620.1}   & 553.8  & 3410.4 \\
            Road Runner     & 7845.0    & 11.5      & 11843.4     & 2606.4     & 5006.1     & 8724.7     & 12213.1 & \textbf{13429.4} & 12337.0  & 12049.7 \\
            Seaquest        & 42054.7   & 68.4      & 304.6      & 272.9      & 315.2     & 310.5      & 558.1    & 532.9   & 471.9  & \textbf{628.3} \\
            Up N Down       & 11693.2   & 533.4     & 3075.0     & 2331.7     & 2646.4     & 3619.1     & \textbf{10859.2} & 10225.2 & 4112.8  & 6675.7 \\
            \midrule
            Interquartile Mean      & 1.000   & 0.000         & 0.183   & 0.117   & 0.113   & 0.224   & 0.337   & 0.374  & 0.292  & \textbf{0.432}\\
            Optimality Gap      & 0.000  & 1.000         & 0.698   & 0.819   & 0.768   & 0.692   & 0.577   & 0.558   & 0.614  & \textbf{0.522}\\
            % Median      & 1.0   & 0         & 14.4  & 16.1  & 20.4  & 17.5  & 26.8  & 41.5  & \textbf{47.2}&  & \\
            % Mean      & 1.0   & 0         & 14.4  & 16.1  & 20.4  & 17.5  & 26.8  & 41.5  & \textbf{47.2}&  & \\
            \bottomrule
        \end{tabular}
    }
    % \vspace{-3pt}   % control the gap between the table bottom and next paragraph
\end{table*}

% To further observe the performance distribution across the games, we use xxx. 

% \begin{wraptable}{r}{7.cm}
\begin{table*}[h]
    % \footnotesize % control font size
    % \scriptsize
    \centering
    \caption{Standard deviations (STDs) of \textit{Baseline} and \ourname~on Atari-100k. The STDs are calculated based on 3 random seeds.}
    \label{table:atari_std}
    % \vskip 0.05in    % control the gap between caption and table
    % \setlength{\tabcolsep}{2.0mm}{
    % \resizebox{\textwidth}{!}{
    \scalebox{0.78}{
        \begin{tabular}{l c c l c c l c c}
            \toprule
            \textbf{Game} & \textbf{Baseline} & \textbf{\ourname} & \textbf{Game} & \textbf{Baseline} & \textbf{\ourname} & \textbf{Game} & \textbf{Baseline} & \textbf{\ourname} \\
            \midrule
            Alien & 61.2 & 79.2 & Crazy Climber & 12980.1 & 2334.5              & Kung Fu Master & 5026.9  & 971.8   \\
            Amidar & 70.7 & 48.0 & Demon Attack & 89.8 & 149.7                 & Ms Pacman & 124.3  & 249.4  \\
            Assault & 4.3 & 28.0 & Freeway & 0.3 & 1.0                      & Pong & 11.5  & 3.1   \\
            Asterix & 32.1 & 43.3 & Frostbite & 1295.3 & 607.3               & Private Eye & 11.5  & 0.0  \\
            Bank Heist & 39.2 & 49.7 & Gopher & 7.9 &  266.0                  & Qbert & 346.6  & 96.3  \\
            Battle Zone & 2070.2 & 1139.6 &  Hero & 2562.3  & 692.0              & Road Runner & 4632.6  & 669.1 \\
            Boxing & 7.3 & 7.4 &  Jamesbond & 108.2  & 20.8                     & Seaquest & 107.1  & 92.2 \\
            Breakout & 1.8 & 1.4 & Kangaroo & 3801.2  & 3025.9                   & Up N Down & 919.3  & 531.5 \\
            Chopper Command & 119.9 & 242.1 & Krull & 605.9  & 792.0  &&& \\
            \bottomrule
        \end{tabular}
    }
    % \vspace{0pt}   % control the gap between the table bottom and next paragraph
% \end{wraptable}
\end{table*}

\subsection{Extended Ablation Study} \label{appendix_ablation}
We give more details of the ablation study. Again, unless otherwise specified, we conduct the ablations on DMControl-100k with 5 random seeds. 

\begin{figure*}[t]%[h] or [t]
	\begin{center}
	\scalebox{0.9}{
		\includegraphics[scale=0.65]{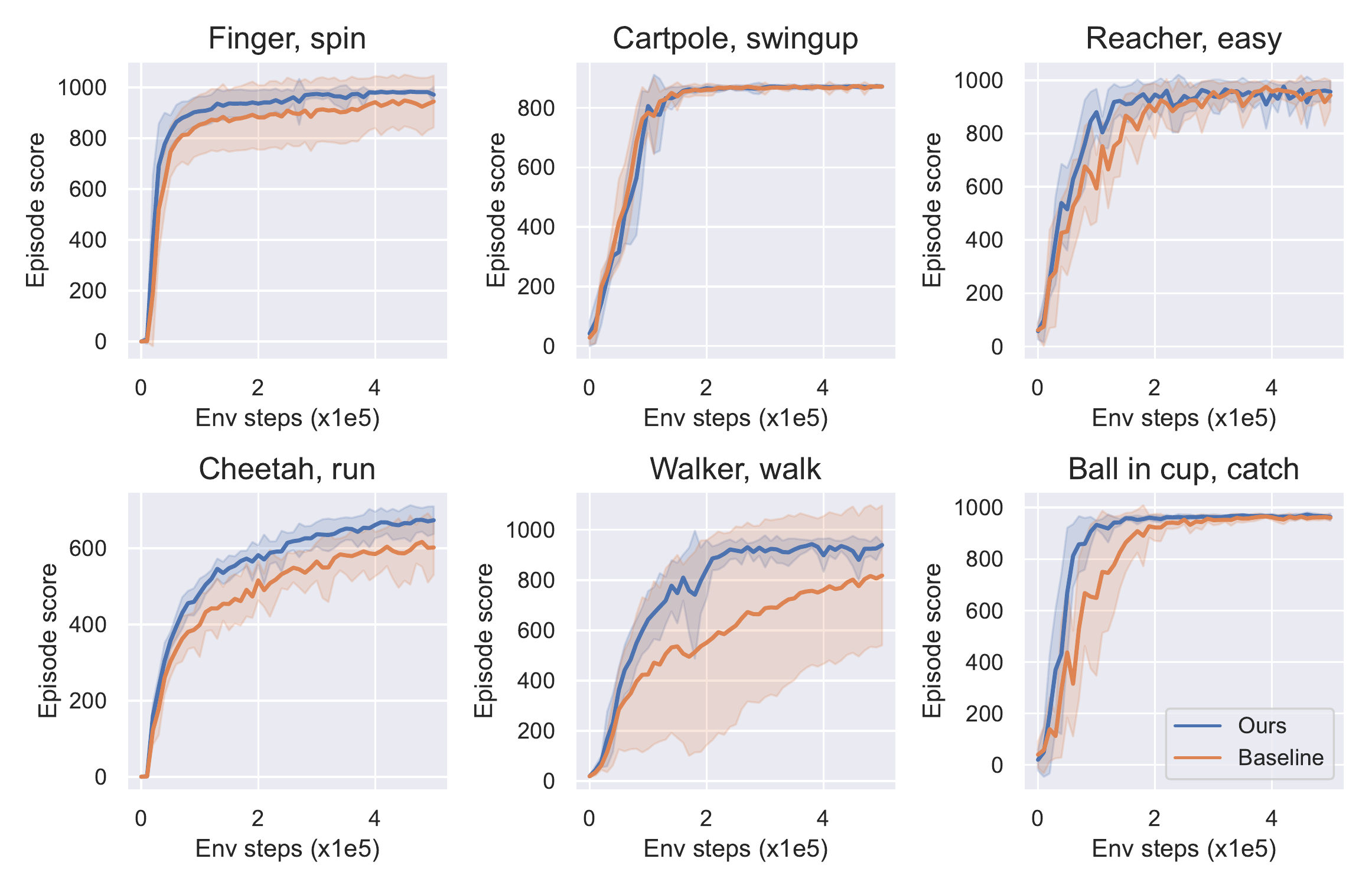} 
	}
	\end{center}
% 	\vspace{-5mm}
	\caption{Test performance during the training period (500k environment steps). Lines denote the mean scores over 10 random seeds, and the shadows are the corresponding standard deviations. In most environments on DMControl, our results (blue lines) are consistently better than \textit{Baseline} (orange lines).
	}
	\label{fig:test_curves}
% 	\vspace{-2mm}
\end{figure*}

\textbf{Effectiveness evaluation.}
Besides the numerical results in Table \ref{table:dmc_compare} in the main manuscript, we present the test score curves during the training process in Figure \ref{fig:test_curves}. Each curve is drawn based on 10 random seeds. The curves demonstrate the effectiveness of the proposed MLR objective. Besides, we observe that MLR achieves more significant gains on the relatively challenging tasks (\egno, \textit{Walker-walk} and \textit{Cheetah-run} with six control dimensions) than on the easy tasks (\egno, \textit{Cartpole-swingup} with a single control dimension). This is because solving challenging tasks often needs more effective representation learning, leaving more room for MLR to play its role.
% \textbf{\textbf{Test performance curve.}}

% \textbf{Decoder depth.} We analyze the influence of using Transformer-based latent decoders with different depths. We show the experimental results in Table \ref{table:ablate_decoder}. Generally, deeper latent decoders lead to worse sample efficiency with lower mean and median scores.
% % Notably, compared to the encoder (4.04M parameters), our decoder is lightweight (40.8K parameters). 
% Similar to the designs in \cite{he2022masked,xie2022simmim}, it is appropriate to use a lightweight decoder in \ourname, because we expect the predicting masked information to be mainly completed by the encoder instead of the decoder. Note that the state representations inferred by the encoder are adopted for the policy learning in RL. 

\textbf{Similarity loss.} MLR performs the prediction in the latent space where the value range of the features is unbounded. Using cosine similarity enables the optimization to be conducted in a normalized space, which is more robust to outliers. We compare MLR models using mean squared error (MSE) loss and cosine similarity loss in Table \ref{table:ablate_metric_heads}.
We find that using MSE loss is worse than using cosine similarity loss. Similar observations are found in SPR \cite{schwarzer2021dataefficient} and BYOL \cite{grill2020byol}.

\textbf{Projection and prediction heads.} MLR adopt the projection and prediction heads following the widely used design in prior works \cite{grill2020byol,schwarzer2021dataefficient}. We study the impact of the heads in Table \ref{table:ablate_metric_heads}. The results show that using both projection and prediction heads performs best, which is consistent with the observations in the aforementioned prior works.

% \begin{wraptable}{r}{0.5\textwidth}
\begin{table*}[t]
    % \vspace{-15pt}
    \footnotesize % control font size
    \centering
    \caption{Ablation study of similarity loss (\textit{SimLoss}), projection (\textit{Proj.}) and prediction (\textit{Pred.}) heads.}
    \label{table:ablate_metric_heads}
    \scalebox{1}{
        \begin{tabular}{c c c c c c}
            \toprule
            \textbf{Model} & \textbf{SimLoss} & \textbf{Proj.} & \textbf{Pred.} & \textbf{Mean} & \textbf{Median}\\ \hline
            \specialrule{0em}{1.5pt}{1pt}   % control space gap
            Baseline & - & - & - & 613.7 & 613.0\\
            \multirow{4}{*}{\ourname} & Cosine & &  & 722.5 & 770.0\\
            & Cosine & & \checkmark & 750.3 & 819.5\\
            & Cosine &\checkmark &\checkmark  & \textbf{767.3} & \textbf{833.0} \\
            & MSE & \checkmark & \checkmark & 704.8 & 721.0\\
            \midrule
        \end{tabular}
    }
    % \vspace{-10pt}
\end{table*}
% \end{wraptable}

\textbf{Sequence length.}
Table \ref{table:ablate_seq_len} shows the results of the observation sequence length $K$ at \{8, 16, 24\}. A large $K$ (\egno, 24) does not bring further performance improvement as the network can reconstruct the missing content in a trivial way like copying and pasting the missing content from other states. In contrast, a small $K$ like 8 may not be sufficient for learning rich context information. A sequence length of 16 is a good trade-off in our experiment.
\begin{table*}[h]
    \vspace{0pt}
    \footnotesize % control font size
    \centering
    \caption{Ablation study of sequence length $K$.}
    \label{table:ablate_seq_len}
    \scalebox{1}{
        \begin{tabular}{l c c c c}
            \toprule
            \textbf{Env.} & \textbf{Baseline} & \textbf{K=8} & \textbf{K=16} & \textbf{K=24} \\ \hline
            \specialrule{0em}{1.5pt}{1pt}   % control space gap
            Finger, spin & 822 $\pm$ 146 & 816 $\pm$ 129   & \textbf{907 $\pm$ 69}     & 875 $\pm$ 63\\
            Cartpole, swingup & 782 $\pm$ 74 & \textbf{857 $\pm$ 3}     & 791 $\pm$ 50      & 781 $\pm$ 58 \\
            Reacher, easy & 557 $\pm$ 137 & 779 $\pm$ 116    & \textbf{875 $\pm$ 92}     & 736 $\pm$ 247 \\
            Cheetah, run & 438 $\pm$ 33 & 469 $\pm$ 51     & \textbf{495 $\pm$ 13}     & 454 $\pm$ 41 \\
            Walker, walk & 414 $\pm$ 310 & 473 $\pm$ 264   & \textbf{597 $\pm$ 102}    & 533 $\pm$ 98 \\
            Ball in cup, catch & 669 $\pm$ 310 & 910 $\pm$ 58   & 939 $\pm$ 9      & \textbf{944 $\pm$ 22} \\ 
            \midrule
            Mean & 613.7 & 717.3 & \textbf{767.3} & 720.5 \\ 
            Median & 613.0 & 797.5 & \textbf{833.0} & 758.5 \\ 
            \midrule
        \end{tabular}
    }
    % \vspace{-5pt}
\end{table*}

\begin{figure*}
     \centering
     \begin{subfigure}[b]{0.49\textwidth}
         \centering
         \includegraphics[width=\textwidth]{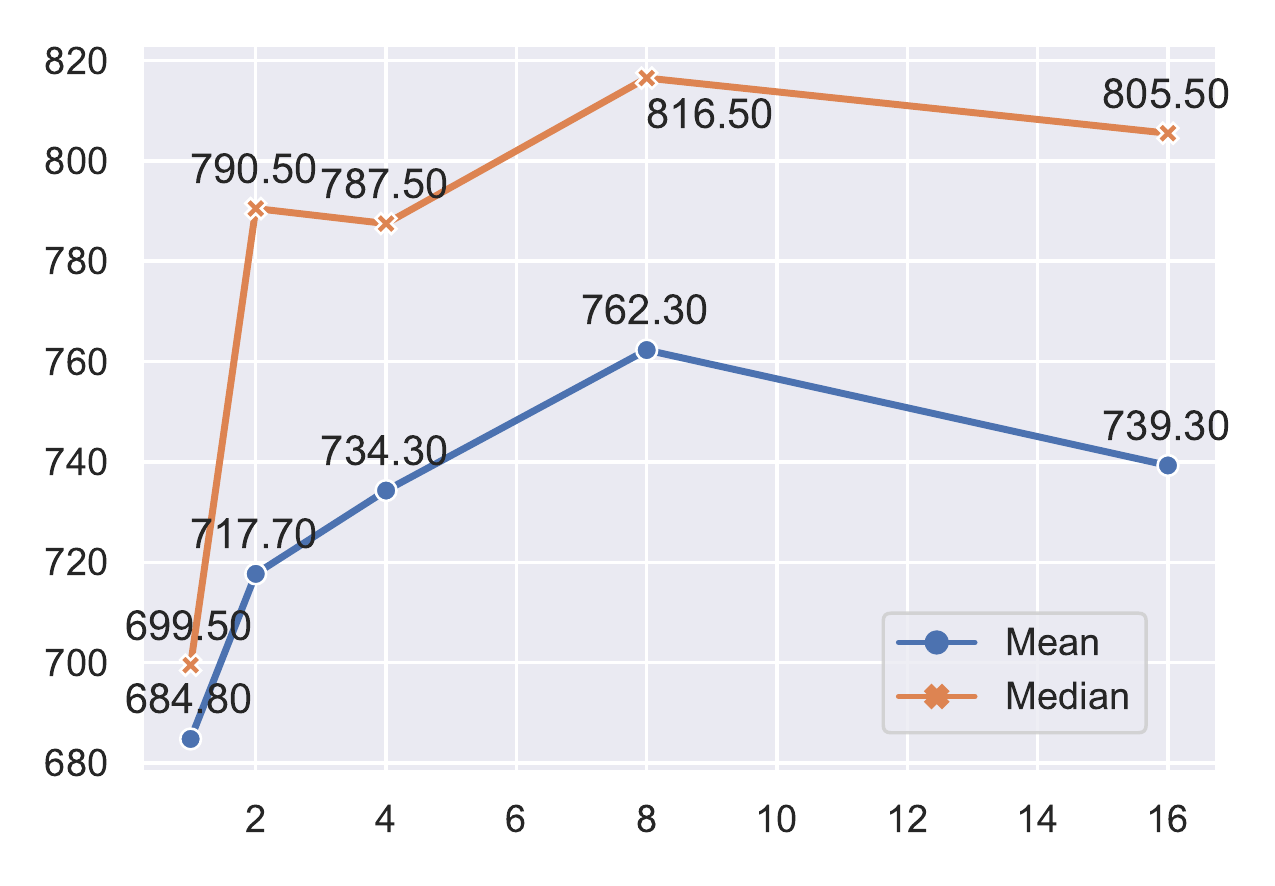}
         \caption{Cube depth $k$}\label{}\label{subfig:ablate_cube_depth}
     \end{subfigure}
     \hfill
     \begin{subfigure}[b]{0.49\textwidth}
         \centering
         \includegraphics[width=\textwidth]{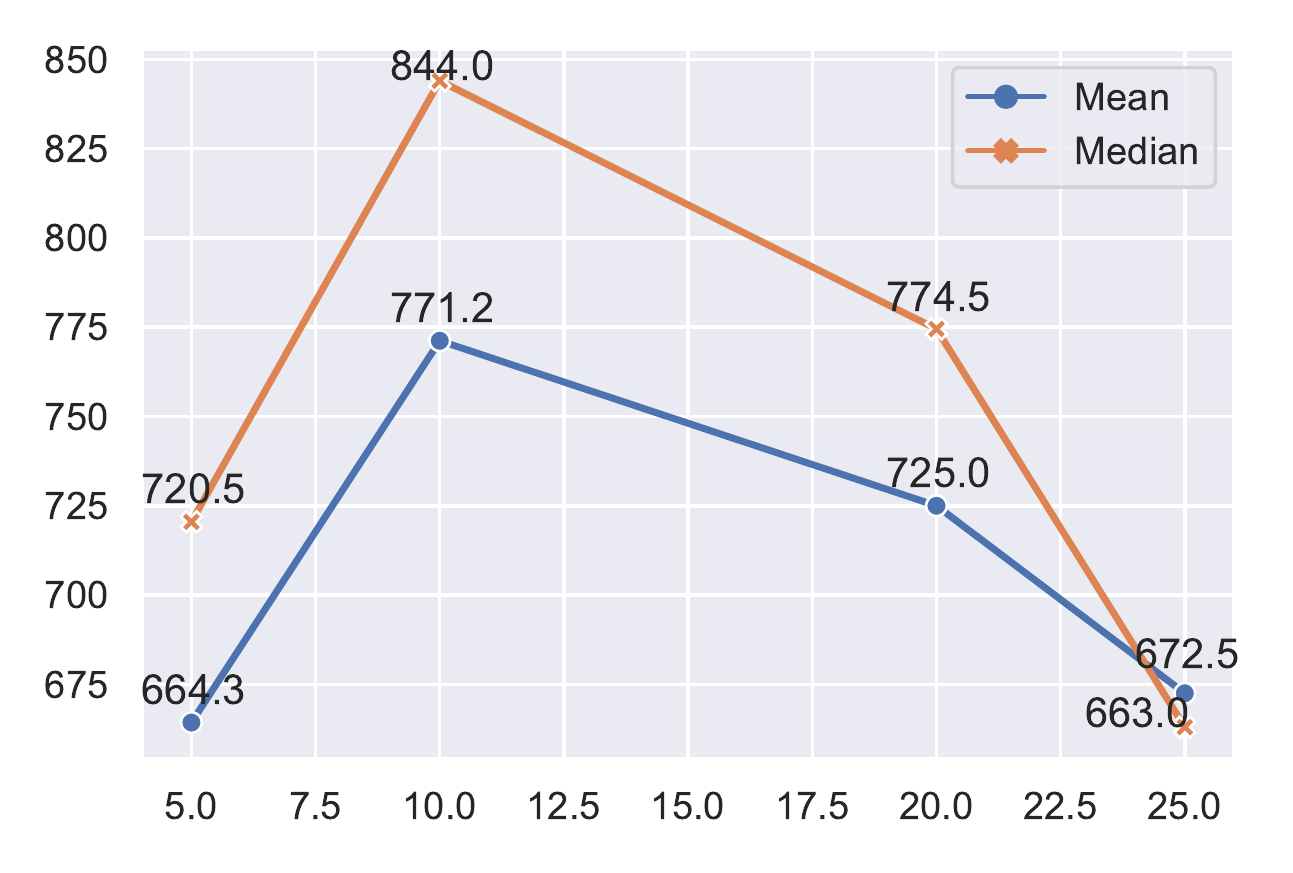}
         \caption{Cube spatial size $h$ \& $w$}\label{}\label{subfig:ablate_patchsize}
     \end{subfigure}
	\caption{Ablation studies of (a) cube depth $k$ and (b) cube spatial size $h$ \& $w$. The result of each model is averaged over 3 random seeds.
	}
	\label{fig:ablate_cube_shape}
\end{figure*}

\textbf{Cube size.}
Our space-time cube can be flexibly designed. We investigate the influence of temporal depth $k$ and the spatial size ($h$ and $w$, $h=w$ by default). The results based on 3 random seeds are shown in Figure \ref{fig:ablate_cube_shape}. In general, a proper cube size leads to good results. The spatial size has a large influence on the final performance. A moderate spatial size (\egno, $10 \times 10$) is good for \ourname~. The performance generally has an upward tendency when increasing the cube depth $k$. However, a cube mask with too large $k$ possibly masks some necessary contents for the reconstruction and hinders the training.

% \subsection{Discussion}

\subsection{Discussion} \label{appendix_discussion}
\textbf{More analysis on MLR-S, MLR-T and MLR.}
\textit{MLR-S} masks spatial patches for each frame independently while \textit{MLR-T} performs masking of entire frames. \textit{MLR-S} and \textit{MLR-T} enable a model to capture rich spatial and temporal contexts, respectively.
We find that some tasks (\egno, \textit{Finger-spin} and \textit{Walker-walk}) have more complicated spatial contents than other tasks on the DMControl benchmarks, requiring stronger spatial modeling capacity. Therefore, \textit{MLR-S} performs better than \textit{MLR-T} on these tasks. While for tasks like \textit{Cartpole-swingup} and \textit{Ball-in-cup-catch}, where objects have large motion dynamics, temporal contexts are more important, and \textit{MLR-T} performs better. \textit{MLR} using space-time masking harnesses both advantages in most cases. But it is slightly inferior to \textit{MLR-S/-T} in \textit{Finger-spin} and \textit{Cartpole-swingup} respectively, due to the sacrifice of full use of spatial or temporal context as in \textit{MLR-S} or \textit{MLR-T}.

\tcr{\textbf{Evaluation on learned representations.} We evaluate the learned representations from two aspects: \textbf{(i) Pretraining evaluation}. We conduct pretraining experiments to test the effectiveness of the learned representation. \textbf{\textit{Data collection}}: We use a 100k-steps pretrained \textit{Baseline} model to collect 100k transitions. \textbf{\textit{MLR pretraining}}: We use the collected data to pretrain MLR without policy learning (\ieno, only with MLR objective). \textbf{\textit{Representation testing}}: We compare two models on DMControl-100k, \textit{Baseline} with the encoder initialized by the MLR-pretrained encoder (denoted as \textit{MLR-Pretraining}) and \textit{Baseline} without pretraining (\ieno, \textit{Baseline}). The results in Table \ref{table:pretraining} show that \textit{MLR-Pretraining} outperforms \textit{Baseline} which is without pretraining but still underperforms MLR which jointly learns the RL objective and the MLR auxiliary objective (also denoted as \textit{MLR-Auxiliary}). This validates the importance of the learned state representations more directly, but is not the best practice to take the natural of RL into account for getting the most out of MLR. This is because that RL agents learn from interactions with environments, where the experienced states vary as the policy network is updated.  \textbf{(ii) Regression accuracy test.} We compute the cosine similarities between the learned representations from the masked observations and those from the corresponding original observations (\ieno, observations without masking). The results in Table \ref{table:correlation} show that there are high cosine similarity scores of the two representations in MLR while low scores in \textit{Baseline}. This indicates that the learned representations of MLR are more predictive and informative.}
% \begin{wraptable}{r}{0.5\textwidth}
\begin{table*}[t]
    % \vspace{-15pt}
    \footnotesize % control font size
    \centering
    \caption{Comparison of non-pretraining \textit{Baseline} (\ieno, \textit{Baseline}), MLR-pretrained \textit{Baseline} (denoted as \textit{MLR-Pretraining}) and our joint learning MLR (denoted as \textit{MLR-Auxiliary}) on DMControl-100k.}
    \label{table:pretraining}
    \scalebox{1}{
        \begin{tabular}{c c c}
            \toprule
            \textbf{DMControl-100k} & \textbf{Cheetah, run} & \textbf{Reacher, easy} \\ \hline
            \specialrule{0em}{1.5pt}{1pt}   % control space gap
            Baseline & 438 $\pm$ 33 & 557 $\pm$ 137\\
            \ourname-Pretraining & 468 $\pm$ 27 & 862 $\pm$ 180 \\
            \ourname-Auxiliary & \textbf{495 $\pm$ 13} & \textbf{875 $\pm$ 92} \\
            \midrule
        \end{tabular}
    }
    % \vspace{-10pt}
\end{table*}
% \end{wraptable}
% \begin{wraptable}{r}{0.5\textwidth}
\begin{table*}[t]
    % \vspace{-15pt}
    \footnotesize % control font size
    \centering
    \caption{Cosine similarities of the learned representations from the masked observations and those from the original observations in \textit{Baseline} and MLR.}
    \label{table:correlation}
    \scalebox{1}{
        \begin{tabular}{c c c}
            \toprule
            \textbf{Cosine Similarity} & \textbf{Cheetah, run} & \textbf{Reacher, easy} \\ \hline
            \specialrule{0em}{1.5pt}{1pt}   % control space gap
            Baseline & 0.366 & 0.254 \\
            \ourname & \textbf{0.930} & \textbf{0.868} \\
            \midrule
        \end{tabular}
    }
    % \vspace{-10pt}
\end{table*}
% \end{wraptable}

\tcr{\textbf{Performance on more challenging control tasks.} We further investigate the effectiveness of MLR on more challenging control tasks such as \textit{Reacher-hard} and \textit{Walker-run}. We show the test scores based on 3 random seeds at 100k and 500k steps in Table \ref{table:challenging_tasks}. Our MLR still significantly outperforms \textit{Baseline}.}

% \begin{wraptable}{r}{0.5\textwidth}
\begin{table*}[t]
    % \vspace{-15pt}
    \footnotesize % control font size
    \centering
    \caption{Comparison of \textit{Baseline} and \ourname~on more challenging DMControl tasks.}
    \label{table:challenging_tasks}
    \scalebox{1}{
        \begin{tabular}{c c c c}
            \toprule
            \textbf{Steps} & \textbf{Model} & \textbf{Reacher, hard} & \textbf{Walker, run} \\ \hline
            \specialrule{0em}{1.5pt}{1pt}   % control space gap
            \multirow{2}{*}{100k} & Baseline & 341 $\pm$ 275 & 105 $\pm$ 47\\
            & \ourname & \textbf{624 $\pm$ 220} & \textbf{181 $\pm$ 19} \\
            \midrule
            % \specialrule{0em}{1.5pt}{1pt}   % control space gap
            \multirow{2}{*}{500k} & Baseline & 669 $\pm$ 290 & 466 $\pm$ 39\\
            & \ourname & \textbf{844 $\pm$ 129} & \textbf{576 $\pm$ 25} \\
            \midrule
        \end{tabular}
    }
    % \vspace{-10pt}
\end{table*}
% \end{wraptable}

\textbf{The relationship between PlayVirtual and our MLR.}
The two works have a consistent purpose, \ieno, improving RL sample efficiency, but address this from two different perspectives. PlayVirtual focuses on how to generate more trajectories for enhancing representation learning. In contrast, our MLR focuses on how to exploit the data more efficiently by promoting the model to be predictive of the spatial and temporal context through masking for learning good representations. They are compatible and have their own specialities, while our MLR outperforms PlayVirtual on average.

\textbf{Application and limitation.} While we adopt the proposed MLR objective to two strong baseline algorithms (\ieno, SAC and Rainbow) in this work, MLR is a general approach for improving sample efficiency in vision-based RL and can be applied to most existing vision-based RL algorithms (\egno, EfficientZero \footnote{EfficientZero augments MuZero \cite{schrittwieser2020mastering} with an auxiliary self-supervised learning objective similar to SPR \cite{schwarzer2021dataefficient} and achieves a strong sample efficiency performance on Atari games. We find that EfficientZero and our MLR have their skilled games on the Atari-100k benchmark. While EfficientZero wins more games than MLR (EfficientZero 18 versus MLR 8), the complexity of EfficientZero is much higher than MLR. Our MLR is a generic auxiliary objective and can be applied to EfficientZero. We leave the application in future work.} \cite{ye2021mastering}). We leave more applications of MLR in future work. We have shown the effectiveness of the proposed MLR on multiple continuous and discrete benchmarks. Nevertheless, there are still some limitations to MLR. \tcr{When we take a closer look at the performance of MLR and \textit{Baseline} on different kinds of Atari games, we find that MLR brings more significant performance improvement on games with backgrounds and viewpoints that do not change drastically (such as \textit{Qbert} and \textit{Frostbite}) than on games with drastically changing backgrounds/viewpoints or vigorously moving objects (such as \textit{Crazy Climber} and \textit{Freeway}). This may be because there are low correlations between adjacent regions in spatial and temporal dimensions on the games like \textit{Crazy Climber} and \textit{Freeway} so that it is more difficult to exploit the spatial and temporal contexts by our proposed mask-based latent reconstruction.
% MLR can bring significant performance improvement on games with backgrounds and viewpoints that do not change drastically such as Qbert and Frostbite, but is not good at games with drastically changing backgrounds/viewpoints or objects of high motion dynamics such as Crazy Climber and Freeway. This may be because masked vision modeling assumes continuity of the original visual signal \cite{cao2022understand}. 
Besides,} MLR requires several hyperparameters (such as mask ratio and cube shape) that might need to be adjusted for particular applications.

\section{Broader Impact} \label{appendix_impact}
Although the presented mask-based latent reconstruction (MLR) should be categorized as research in the field of RL, the concept of reconstructing the masked content in the latent space may inspire new approaches and investigations in not only the RL domain but also the fields of computer vision and natural language processing. MLR is simple yet effective and can be conveniently applied to real-world applications such as robotics and gaming AI. However, specific uses may have positive or negative effects (\ieno, the dual-use problem). We should follow the responsible AI policies and consider safety and ethical issues in the deployments.

% \newpage
% \input{tables/atari_hyperparam}
% \newpage
% \input{tables/dmc_hyperparam}

\newpage
% \begin{wraptable}{r}{7.cm}
\begin{table*}[h]
    % \footnotesize % control font size
    % \scriptsize
    \centering
    \caption{Hyperparameters used for DMControl.}
    \label{table:dmc_hyperparam}
    % \vskip 0.05in    % control the gap between caption and table
    % \setlength{\tabcolsep}{2.0mm}{
    % \resizebox{\textwidth}{!}{
    \scalebox{1.0}{
        \begin{tabular}{l c c c c l}
            \toprule
            \textbf{Hyperparameter} & & & & & \textbf{Value} \\
            \midrule
            Frame stack & & & & & 3 \\
            Observation rendering & & & & & (100, 100) \\
            Observation downsampling & & & & & (84, 84) \\
            % Augmentation for policy learning  & & & & & Random crop \\
            % Augmentation for auxiliary task  & & & & & Random crop and random intensity \\
            Augmentation  & & & & & Random crop and random intensity \\
            Replay buffer size & & & & & 100000 \\
            Initial exploration steps & & & & & 1000 \\
            Action repeat & & & & & 2 \textit{Finger-spin} and \textit{Walker-walk};\\
             & & & & & 8 \textit{Cartpole-swingup}; \\
             & & & & & 4 otherwise \\
            
            Evaluation episodes & & & & & 10 \\
            
            Optimizer & & & & & Adam \\
            % $(\beta_1, \beta_2) \rightarrow (\theta_f, \xi_h, \xi_b, \omega)$ & & & & & (0.9, 0.999) \\
            ~~~~$(\beta_1, \beta_2) \rightarrow (\theta_f, \theta_\phi, \theta_g, \theta_q, \theta_\omega)$ & & & & & (0.9, 0.999) \\
            ~~~~$(\beta_1, \beta_2) \rightarrow (\alpha)$ (temperature in SAC) & & & & & (0.5, 0.999) \\
            Learning rate $(\theta_f, \theta_\omega)$ & & & & & 0.0002 \textit{Cheetah-run} \\
             & & & & & 0.001 otherwise \\
            Learning rate $(\theta_f, \theta_\phi, \theta_g, \theta_q)$ & & & & & 0.0001 \textit{Cheetah-run} \\
             & & & & & 0.0005 otherwise \\
            Learning rate warmup $(\theta_f, \theta_\phi, \theta_g, \theta_q)$ & & & & & 6000 steps \\
            Learning rate $(\alpha)$ & & & & & 0.0001 \\
            Batch size for policy learning & & & & & 512 \\
            Batch size for auxiliary task & & & & & 128 \\
            
            Q-function EMA $m$ & & & & & 0.99 \\
            Critic target update freq & & & & & 2 \\

            Discount factor & & & & & 0.99 \\
            Initial temperature & & & & & 0.1 \\

            Target network update period & & & & & 1 \\
            % Target network EMA $m$ & & & & & 0.05 (expect for 0.1 in Walker, walk) \\
            Target network EMA $m$ & & & & & 0.9 \textit{Walker-walk} \\
             & & & & & 0.95 otherwise \\
            State representation dimension $d$ & & & & & 50 \\
            \midrule
            \textbf{\ourname~Specific Hyperparameters}& & & & & \\
            \midrule
            Weight of \ourname~loss $\lambda$ & & & & & 1 \\
            Mask ratio $\eta$ & & & & & 50\% \\
            Sequence length $K$ & & & & & 16 \\
            Cube spatial size $h \times w$ & & & & & 10 $\times$ 10 \\
            Cube depth $k$ & & & & & 4 \textit{Cartpole-swingup} and \textit{Reacher-easy}\\
            & & & & & 8 otherwise \\
            Decoder depth $L$ (number of attention layers) & & & & & 2 \\
            \bottomrule
        \end{tabular}
    }
    % \vspace{0pt}   % control the gap between the table bottom and next paragraph
% \end{wraptable}
\end{table*}

\newpage
% \begin{wraptable}{r}{7.cm}
\begin{table*}[h]
    % \footnotesize % control font size
    % \scriptsize
    \centering
    \caption{Hyperparameters used for Atari.}
    \label{table:atari_hyperparam}
    % \vskip 0.05in    % control the gap between caption and table
    % \setlength{\tabcolsep}{2.0mm}{
    % \resizebox{\textwidth}{!}{
    \scalebox{1.0}{
        \begin{tabular}{l c c c c l}
            \toprule
            \textbf{Hyperparameter} & & & & & \textbf{Value} \\
            \midrule
            Gray-scaling & & & & & True \\
            Frame stack & & & & & 4 \\
            Observation downsampling & & & & & (84, 84) \\
            % Augmentation for policy learning  & & & & & Random crop and random intensity \\
            % Augmentation for auxiliary task  & & & & & Random crop and random intensity \\
            Augmentation  & & & & & Random crop and random intensity \\
            Action repeat & & & & & 4 \\
            
            Training steps & & & & & 100K \\
            Max frames per episode & & & & & 108K \\
            Reply buffer size & & & & & 100K \\
            Minimum replay size for sampling & & & & & 2000 \\
            Mini-batch size & & & & & 32 \\

            Optimizer & & & & & Adam \\
            Optimizer: learning rate & & & & & 0.0001 \\
            Optimizer: $\beta_1$ & & & & & 0.9 \\
            Optimizer: $\beta_2$ & & & & & 0.999 \\
            Optimizer: $\epsilon$ & & & & & 0.00015 \\
            Max gradient norm & & & & & 10 \\

            Update & & & & & Distributional Q \\
            Dueling & & & & & True \\
            Support of Q-distribution & & & & & 51 bins \\
            Discount factor & & & & & 0.99 \\
            Reward clipping Frame stack & & & & & [-1, 1] \\
            
            Priority exponent & & & & & 0.5 \\
            Priority correction & & & & & 0.4 $\rightarrow$ 1 \\
            Exploration & & & & & Noisy nets \\
            Noisy nets parameter & & & & & 0.5 \\
            
            Evaluation trajectories & & & & & 100 \\
            
            Replay period every & & & & & 1 step \\
            Updates per step & & & & & 2 \\
            Multi-step return length & & & & & 10 \\
            
            Q network: channels & & & & & 32, 64, 64 \\
            Q network: filter size & & & & & 8 $\times$ 8, 4 $\times$ 4, 3 $\times$ 3 \\
            Q network: stride & & & & & 4, 2, 1 \\
            Q network: hidden units & & & & & 256 \\
            
            Target network update period & & & & & 1 \\
            $\tau$ (EMA coefficient) & & & & & 0 \\
            \midrule
            \textbf{\ourname~Specific Hyperparameters}& & & & & \\
            \midrule
            Weight of \ourname~loss $\lambda$ & & & & & 5 \textit{Pong} and \textit{Up N Down} \\
            & & & & & 1 otherwise \\
            Mask ratio $\eta$ & & & & & 50\% \\
            Sequence length $K$ & & & & & 16 \\
            Cube spatial size $h \times w$ & & & & & 12 $\times$ 12 \\
            Cube depth $k$ & & & & & 8 \\
            Decoder depth $L$ (number of attention layers) & & & & & 2 \\
            \bottomrule
        \end{tabular}
    }
    % \vspace{0pt}   % control the gap between the table bottom and next paragraph
% \end{wraptable}
\end{table*}

            % Q network: channels & & & & & 32, 64, 64 \\
            % Q network: filter size & & & & & 8 $\times$ 8, 4 $\times$ 4, 3 $\times$ 3 \\
            % Q network: stride & & & & & 4, 2, 1 \\
            % Q network: hidden units & & & & & 256 \\
            % DM network: channels & & & & & 64, 64 \\
            % DM network: filter size & & & & & 64, 64 \\
            % RDM network: channels & & & & & 64, 64 \\
            % Non-linearity & & & & & ReLU \\

\end{document}